\documentclass[a4paper,11pt]{article}

\usepackage{geometry}
\usepackage{graphicx}
\usepackage[x11names]{xcolor}
\usepackage{amsmath}
\usepackage[colorlinks,citecolor=Green3]{hyperref}
\usepackage{booktabs}
\usepackage{diagbox}
\usepackage{multirow}
\newcommand{\tabincell}[2]{\begin{tabular}{@{}#1@{}}#2\end{tabular}}

\title{\textbf{A Robust Real-Time Computing-based Environment Sensing System for Intelligent Vehicle}}
\author{Qiwei Xie$^{1,\dagger}$, Qian Long$^{2,\dagger}$, Liming Zhang$^{3,\star}$, Zhao Sun$^4$ \\
1. Data Mining Laboratory, Beijing University of Technology, Beijing 100124, China; \\
Institute of Automation, Chinese Academy of Sciences, Beijing 100190, China. \\
2. Yunan Observatories, Chinese Academy of Sciences, Kunming, China. \\
3. Faculty of Science and Technology, University of Macau, Macau, China. \\
4. Faculty of Information Technology, Macau University of Science and Technology, Macau, China.}
\date{}

\begin{document}
\maketitle

\begin{abstract}
For intelligent vehicles, sensing the 3D environment is the first but crucial step. In this paper, we build a real-time advanced driver assistance system based on a low-power mobile platform. The system is a real-time multi-scheme integrated innovation system, which combines stereo matching algorithm with machine learning based obstacle detection approach and takes advantage of the distributed computing technology of a mobile platform with GPU and CPUs. First of all, a multi-scale fast MPV (Multi-Path-Viterbi) stereo matching algorithm is proposed, which can generate robust and accurate disparity map. Then a machine learning, which is based on fusion technology of monocular and binocular, is applied to detect the obstacles. We also advance an automatic fast calibration mechanism based on Zhang's calibration method. Finally, the distributed computing and reasonable data flow programming are applied to ensure the operational efficiency of the system. The experimental results show that the system can achieve robust and accurate real-time environment perception for intelligent vehicles, which can be directly used in the commercial real-time intelligent driving applications.

\end{abstract}

\section{Introduction}
For autonomous driving or advanced driver assistance systems (ADAS), Object detection based on 3D environment perception is one of the key components. There are usually four major types of 3D environment perception approaches in the literature, including stereo-vision-based, LiDAR-based, radar-based, and multi-sensor-based hybrid approaches \cite{zhu2017overview,long2014real,hata2015feature,xie2018pixels,gao2018object}.

LiDAR and radar have been applied extensively to detect obstacles in intelligent vehicles. They use laser light or signal of radio waves to detect the distance to objects, respectively \cite{zhao2018key,van2018autonomous,zhang2016study,montemerlo2008junior,kammel2008team}.
Compared to LIDAR and radar, binocular stereo vision does not involve motion artifacts and can generate much denser depth information. Besides, neither LiDAR nor radar can detect objects smaller than a certain size. For example, for a typical LIDAR-Velodyne HDL-$32$E, we can easily calculate that the minimum detectable height of the object is $0.05$m where the LIDAR is mounted $2$m from the ground. In other words, this system may miss objects below $0.05$m. However, for vehicles traveling on highways, objects of the above dimension may become a potential threat to safe driving. The binocular stereo matching algorithm can generate dense enough disparity information for detecting this kind of small objects due to its much higher image resolution compared to LIDAR and radar.

This paper presents a real-time ADAS based on the binocular stereo vision running on a low-power mobile platform. A fast calibration system is proposed to achieve calibration of the binocular camera. A multi-scale fast MPV (multi-path-viterbi) algorithm is also proposed to adapt to limited resources of the mobile platform. Real-time detection is achieved by using a simple and effective recognition scheme. In addition, distributed computing technology of processing units is advanced to speed up computing and enhance the robustness of the system. The extensive experiments verify the feasibility of the auxiliary driving system based on binocular and demonstrate the applicability from the perspective of hardware device.

\subsection{Related Works}
We conducted the literature review in four aspects, including fast calibration method, perception system, road detection, and distributed computing technology.

\subsubsection{Fast calibration method.} Binocular camera calibration is the process of estimating the intrinsic parameters of two monocular cameras and the extrinsic parameters of the binocular cameras. In traditional methods, calibration algorithms are used to construct the geometric model of optical lens \cite{weng1992camera,scharstein2002taxonomy,zhang2000flexible}. Recently, calibration methods based on neural network \cite{smith2005automatic} have been proposed. Existing study \cite{zhao2011camera} shows that Zhang?．s technique not only avoids complex operation of traditional calibration methods, but also can ensure higher accuracy and stability. However, Zhang?．s algorithm requires multiple images of a planar calibration grid. Based on Zhang?．s algorithm, we propose a new automatic fast calibration system.

\subsubsection{Perception system.} The main hardware of our perception system is a stereo vision sensor consisting of binocular camera and IPS chip. We adopt the multi-scale fast MPV stereo matching algorithm to implement the stereo matching, which is based on a hierarchical bi-direction Viterbi process constrained by Total Variation (TV) \cite{long2014real, dey2006richardson}. We suggest this strategies  with the specific mathematical proof to exhibit that the proposed fast MPV method is suitable for running at our mobile platform. We also improve the original fast MPV algorithm to a multi-scale fast MPV algorithm by the way of virtual nodes.

\subsubsection{Road detection.} There are several road detection approaches for intelligent vehicles in the literature. They can be divided into three categories, including monocular-based method \cite{kong2010general}, homograph-transformation-based method \cite{guo2012robust} and the stereo-matching-based method. The stereo-matching-based method can be further classified into two sub-categories based on the space where the processing is performed, including v-disparity space \cite{hu2005uv} and and Euclidean space \cite{nedevschi2004high}. The former method is faster than the latter but requires higher road flatness \cite{sappa2007road}.

This paper proposes a fully unsupervised way to detect the object by adding an extra Viterbi process on v-disparity space. We use a fast MPV stereo matching algorithm to generate robust and accurate disparity map compared to other state-of-the-art real-time stereo matching algorithms. This multi-scale fast MPV algorithm includes two parts: estimating disparity and estimate epipolar line distortion. Firstly, we conduct Viterbi process at 4 bi-directional paths \cite{forney1973viterbi, xie2017integration} to estimate disparity. Then, based on the results of Viterbi, a convex optimization equation is derived to estimate epipolar line distortion. Finally, two parts are combined into an online framework to do stereo matching.

Based on the result of binocular stereo matching, we also propose an image fusion technique to combine two images of binocular disparity map and monocular gray map. This fusion method will provide the region of interest (ROI) in images for our system. Finally, we propose a target recognition method based on machine learning to improve the robustness and accuracy for detecting distant objects.

\subsubsection{Processing unit.} In this paper, we propose a hybrid programming method based on the combination of CPU and GPU that is different from other approaches in the literature \cite{power2013heterogeneous}. In our approach the GPU is used to accelerate the CPU's calculations. We take advantage of the distributed computing technology by arranging GPU to run the binocular stereo matching algorithm and CPU to achieve target recognition by using a cascade AdaBoost machine learning method \cite{fleuret2004fast}. The AdaBoost algorithm uses a non-random combination of several weak classifiers and weights to each weak classifier to build a strong classifier with superior performance. In this situation, the system accuracy can be improved.

CUDA is also used in our system, which is a parallel programming model and software environment launched by NVIDIA \cite{vineet2008cuda}. It uses a high-level language as a programming language and provides a large number of high-performance computing instructions. CUDA can fully utilize the GPU's large-scale parallel computing capability \cite{kirk2007nvidia}.

\subsection{Our Contributions}
The contributions of this paper are as follows.
\begin{itemize}
  \item An advanced driver assistance system on mobile platforms for intelligent vehicles is proposed in this paper, which consists of the hardware design (system design and fast calibration device) and software design (algorithms and data management). The system can not only be implemented in real-time, but also be robust under different weather and lighting conditions.

  \item A fast binocular calibration method has been proposed for our system and special fast calibration apparatus is designed for this purpose.

  \item A multi-scale fast MPV algorithm is proposed in order to run a real-time system on mobile platforms. This method can provide dense disparity information, road detection and segmentation, and obstacle detection.

  \item A machine learning classifier is trained in our object recognition system, in which the corresponding ROI sets are labeled artificially. Our classifier is sensitive to distant objects, thus improving their system detection accuracy.

  \item Distributed computing and efficient data management provide an important foundation for the real-time operation of the system.
\end{itemize}

\subsection{Organization of The Paper}
The rest of the paper is organized as follows. The overview of the proposed system is introduced in Section 2. The principle of the proposed system is elaborated in Section 3. The experiment results and analysis are presented in Section 4. Conclusions are drawn in Section 5.

\section{Overview of The Proposed System}
In this section, we briefly introduce the whole system design, multi-scale fast MPV algorithm for stereo matching, the target recognition method based on monocular and binocular fusion, fast calibration method and the device, the hardware structure of the mobile platform, and the data management.

\subsection{System Design}
Our system is designed according to the following procedures, as shown in Figure \ref{working_flow}. First, two images (left and right images) are inputted into the system. The stereo matching algorithm is used to calculate the disparity map, which contains a structural similarity (SSIM) algorithm and a bi-directional Viterbi algorithm. Based on the disparity map, we calculate the histogram of the disparity in the horizontal direction and vertical direction to obtain the road model. Then based on the road model and disparity map, the obstacle ROI is computed. By combining the obstacle ROI and left image, the fusion detection of monocular and binocular can be achieved. Finally, we collect all the ROI sets and label them artificially to train an AdaBoost classifier. The AdaBoost classifier can identify whether there is an obstacle in the ROI. If an obstacle can be detected by the classifier, a red rectangle is marked on the obstacle in the original left image.

\begin{figure}[!h]
	\centering
	\includegraphics[width=14cm]{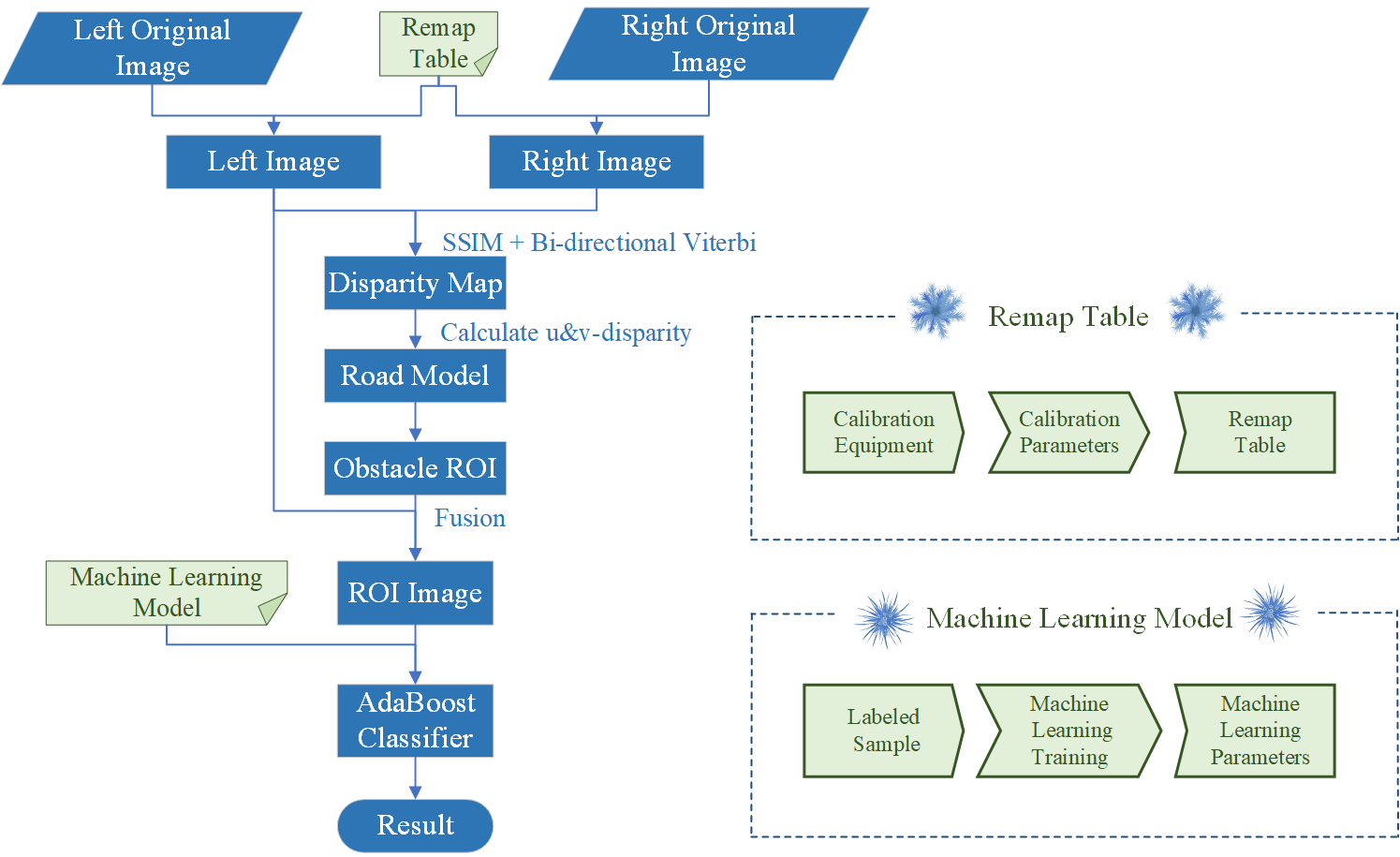}
	\caption{System working flow.}
	\label{working-flow}
\end{figure}

\subsection{Stereo Matching}
Our proposed disparity estimation method has the following characteristics.
\begin{itemize}
  \item A bi-directional Viterbi algorithm for total 4 paths is used to decode the matching cost space and a hierarchical strategy is proposed to merge the 4 paths to further decrease the decoding error in \cite{bahl1974optimal}.

  \item We apply the TV constraint \cite{chambolle2004algorithm} into the Viterbi path in order to approximately model 3D plane at different orientations to achieve similar effects to Total Generalized Variation (TGV) \cite{ranftl2013minimizing} and Slanted-plane models \cite{birchfield1999multiway}.

  \item We apply a fast calculation technique to find the best Viterbi path, and use a multi-scale method to greatly increase the computational speed with a small loss of accuracy.

  \item We use SSIM to measure the pixel difference between left and right images at epiploic lines.
\end{itemize} 

\subsection{Target Recognition}
Based on the result of binocular stereo matching, we propose an image fusion technique to overlay two image products, namely binocular disparity map and monocular gray map \cite{liu2011objective, garcia2017sensor}. This fusion method can provide the ROI of images for our system.

We also propose a target recognition method using machine learning algorithm to improve the robustness and accuracy in detecting distant objects, which has the following characteristics.
\begin{itemize}
  \item Based on the result of stereo matching, we locate the ROI of suspected distant objects in the disparity map, and then extract the corresponding ROI in the monocular image. This processing can reduce the dimension of input data and save the time in window transformation so that it ensures real time detection.

  \item The distant objects are detected by a cascade classifier that is trained by the AdaBoost algorithm based on Local Binary Pattern (LBP) features. These features can measure and describe image local texture information with low sensitivity to illumination. In addition, the algorithm is not complicated and easy to implement. Experiments show that our method is robust and accurate on processors with relatively low performance.
\end{itemize}

\subsection{Fast Calibration}
According to Zhang?．s algorithm, we design an automatic fast calibration method and dedicated device for the system, which allows our system to be calibrated without excessive manual intervention and to ensure consistent calibration results.

\subsection{Hardware Structure of Mobile Platform}
\begin{figure}[!h]
	\centering
	\includegraphics[width=14cm]{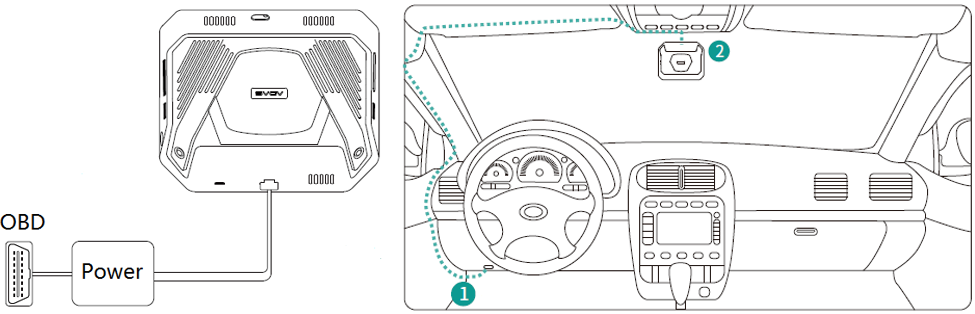}
	\caption{Assembly sketch. Left: System Appearance. Right: installation diagram. 1. OBD power. 2. Our system.}
	\label{assembly_position}
\end{figure}

\begin{figure}[!h]
	\centering
	\includegraphics[width=14cm]{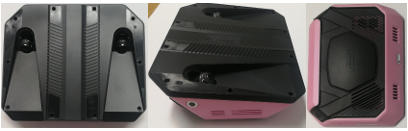}
	\caption{Development platform.}
	\label{Development_platform}
\end{figure}

The system consists of binocular cameras, ISP (Image Signal Processing) module, CPUs \& GPU modules, and power system. The assembly sketch is shown in Figure \ref{assembly_position}. The baseline of our binocular cameras is $120$ mm, the focal length of the lens is $8$ mm and the resolution of the sensors is $640 \times 480$ pixels. Specific parameters of our system are shown in Table \ref{system_params} and our development platform is shown in Figure \ref{Development_platform}.

\begin{table}
	\centering	
	\caption{System parameters.}
	\begin{tabular}[!h]{c|c}		
		\toprule
		Items & Parameters \\
		\midrule
		Effective measurement distance & $3\sim60$ m \\
        Horizontal field of view & $40$ degree \\
        Dynamic range & $120$ DB \\
        Resolution ratio & $640 \times 480$ \\
        Baseline distance & $120$ mm \\
        Focal length & $8$ mm \\
        Data depth & $8$ bits \\
        Service voltage & $12$ V \\
		\bottomrule	
	\end{tabular}
    \label{system_params}	
\end{table}

\begin{figure}[!h]
	\centering
	\includegraphics[width=14cm]{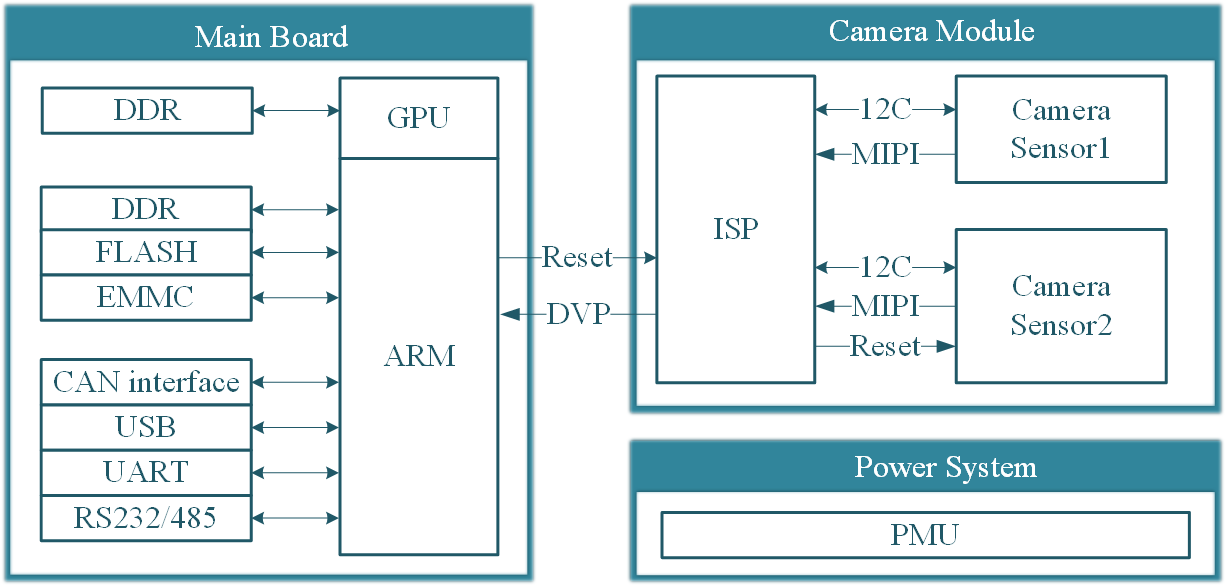}
	\caption{System structure.}
	\label{system_structure}
\end{figure}

In order to improve the efficiency of the operation, parallel computing is adopted in our system. This system is built with an NVIDIA Kepler "GK20a" GPU with 192 SM3.2 CUDA cores (upto 326 GFLOPS) and an NVIDIA "4-Plus-1" 2.32GHz ARM quad-core Cortex-A15 CPU with Cortex-A15 battery-saving shadow-core. Our system structure is shown in Figure \ref{system_structure}, which consists of two cameras and a data processing and control unit. It can provide hundreds of millions of transistors to the processor by Moore law, but most of these transistors are used to make caches in CPU designed to run a single thread program. This can control the processor power consumption within a reasonable range. However, it hinders the further improvement of the performance.

Different from CPU, GPU has a large number of execution and operation units, and only a small amount of data is used for data caching and instruction flow control. So, our combination of CPU and GPU can provide better performance.

\subsection{Data Management}
\begin{figure}[!h]
	\centering
	\includegraphics[width=14cm]{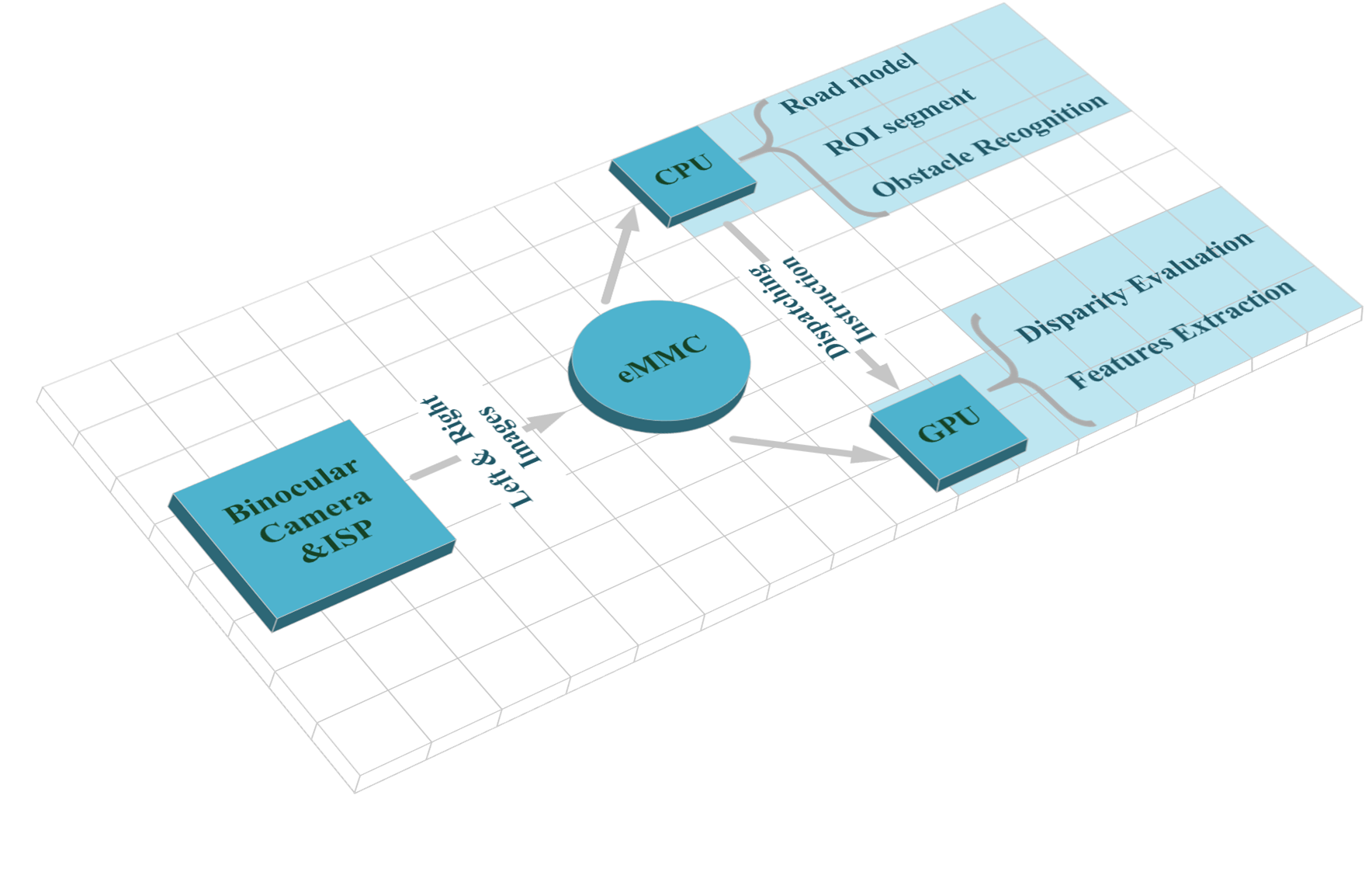}
	\caption{The logical architecture of CPUs and GPU.}
	\label{cpu_gpu}
\end{figure}

As a real-time system, we use distributed computing and reference to robust software partitioning techniques \cite{spacey2012robust} to speed up the operations. In our system, the GPU is only used to run the fast stereo matching algorithm. Here, we refer to \cite{boschetti2016using} programming with CUDA, which implements a higher abstraction model, while the other four CPUs execute instruction scheduling, road detection, target recognition, and other computing tasks, respectively. We refer to \cite{date2019level} using parallel computing methods to assign tasks to the CPU. The main CPU executes instruction scheduling, while performing road detection and target recognition on other three CPUs at same time, as shown in Figure \ref{cpu_gpu}.

\section{The Principles of The Proposed System}
In this section, we present the detailed techniques of our proposed system in eight subsections. The fast-automatic stereo calibration system is first introduced in subsection 3.1. Then, the principle of the stereo matching algorithm, the fast calculation method for finding the best Viterbi path, and the multi-scale image matching approach are presented in subsection 3.2, 3.3, and 3.4, respectively. Next, subsection 3.5, 3.6, and 3.7 respectively describe the road and obstacle detection method, the target recognition, and the distributed computing technology. Finally, the data flow of our system is presented in subsection 3.8.

\subsection{Fast Automatic Stereo Calibration System}
Stereo calibration provides the basis for matching binocular cameras \cite{konolige1998small}. Camera calibration is a prerequisite for the binocular stereo vision system, and the accuracy of the calibration parameters plays a crucial role in subsequent processing in the system \cite{li2018binocular}.  In this paper, our system is calibrated based on Zhang's method \cite{zhang2000flexible}.

Binocular stereo vision works as follows: it uses two imaging devices to acquire two images of the object being measured from different locations, and then calculates the camera?．s internal and external parameters. The distance between the object and the imaging devices can be obtained by calculating the positional deviation between the corresponding pixels of the two images (referred to as ``disparity''). Thereby, three-dimensional information of an object in the camera coordinate system can be acquired. A binocular stereoscopic vision imaging apparatus typically consists of two identical cameras placed side-by-side and spaced apart from one another (called "baseline distance"), commonly referred to as a binocular stereo camera (short as ``binocular camera'' in this paper).

In order to accurately obtain the three-dimensional information of the object to be measured in the camera coordinate system, the binocular camera is usually calibrated according to the "Zhang's camera calibration method" using a black and white grid calibration board, thereby obtaining a series of calibration parameters. These parameters include internal parameter matrices (such as distortion parameter matrix), external parameter matrices, and so on.

\begin{figure}[!h]
	\centering
	\includegraphics[width=14cm]{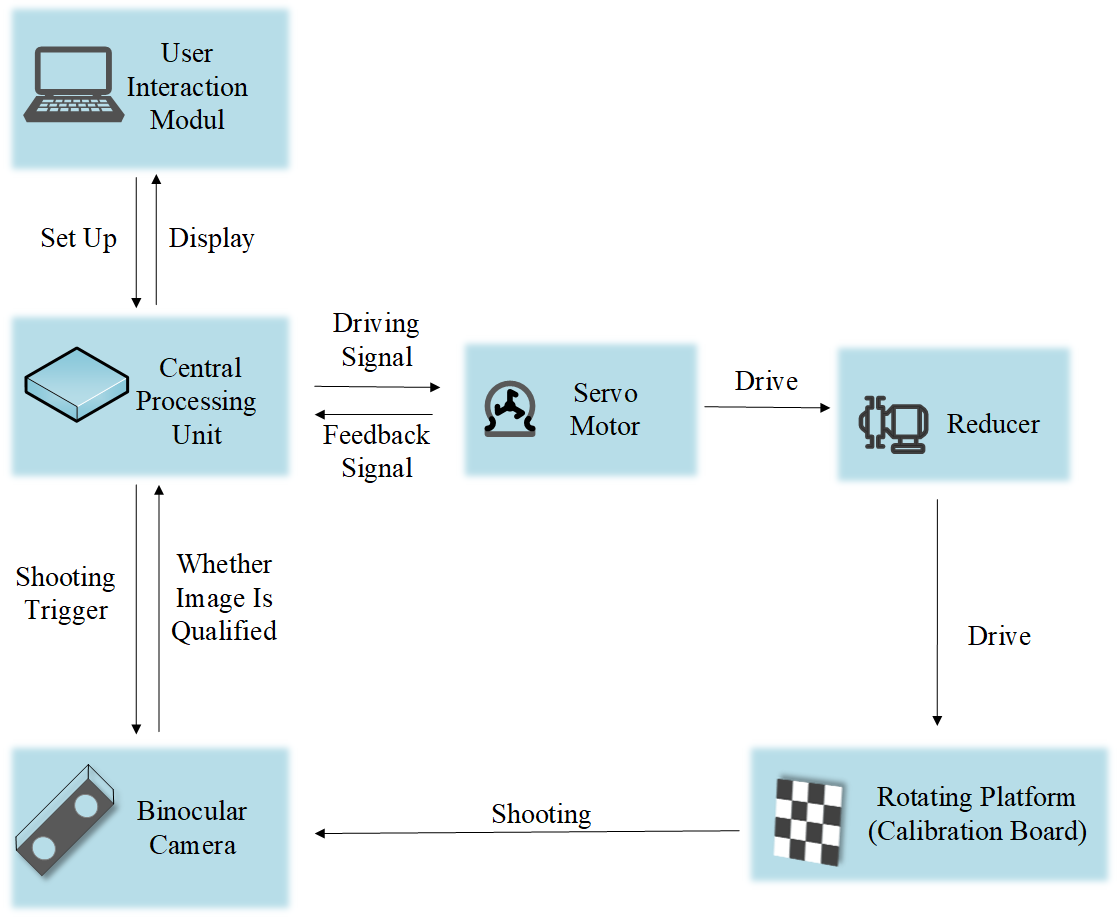}
	\caption{The calibration process of the binocular camera.}
	\label{calib_process}
\end{figure}

The calibration process is as follows: image pairs are captured by cameras $1$ and $2$, respectively. After capturing each image pair, the calibration board is regularly adjusted relative to the position of the binocular camera. In this situation, the binocular camera will acquire dozens of image pairs or even dozens of calibration boards. Each set of image pairs should also meet certain conditions. For example, all the grid angles on the calibration board are located in the field of view of camera 1 and camera 2 and can be accurately extracted by the algorithm. Besides, the calibration board is always flat and clean. The calibration process is shown in Figure \ref{calib_process}.

Since the calibration process of the binocular camera is complicated and the steps are cumbersome, and it is usually operated manually, it becomes the biggest bottleneck that affects the production efficiency in the mass production process of the binocular camera. The proposed fast automatic binocular camera calibration device (hereinafter referred to as ``calibration device'') solves the efficiency problem of the binocular camera calibration process and reduces the calibration time of a single binocular camera to less than $1$ minute.

\begin{figure}[!h]
	\centering
	\includegraphics[width=14cm]{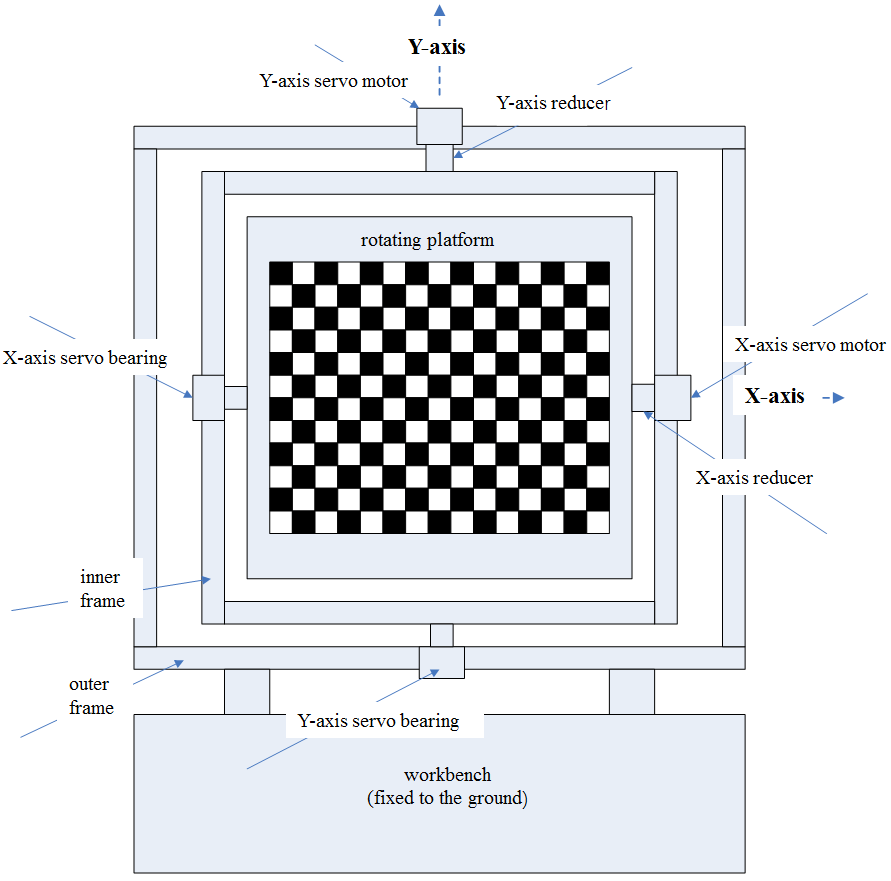}
	\caption{Workbench of binocular camera calibration.}
	\label{calib_device}
\end{figure}

The calibration workbench is shown in Figure \ref{calib_device}. The checkerboard is designed as the calibration board. Through stereo calibration, we obtain the intrinsic and extrinsic parameters and coefficient of the distortion model. We can then calculate a remapping table of the images that indicate the mapping between the pixels in the corrected image and the original image.

\begin{figure}[!h]
	\centering
	\includegraphics[width=14cm]{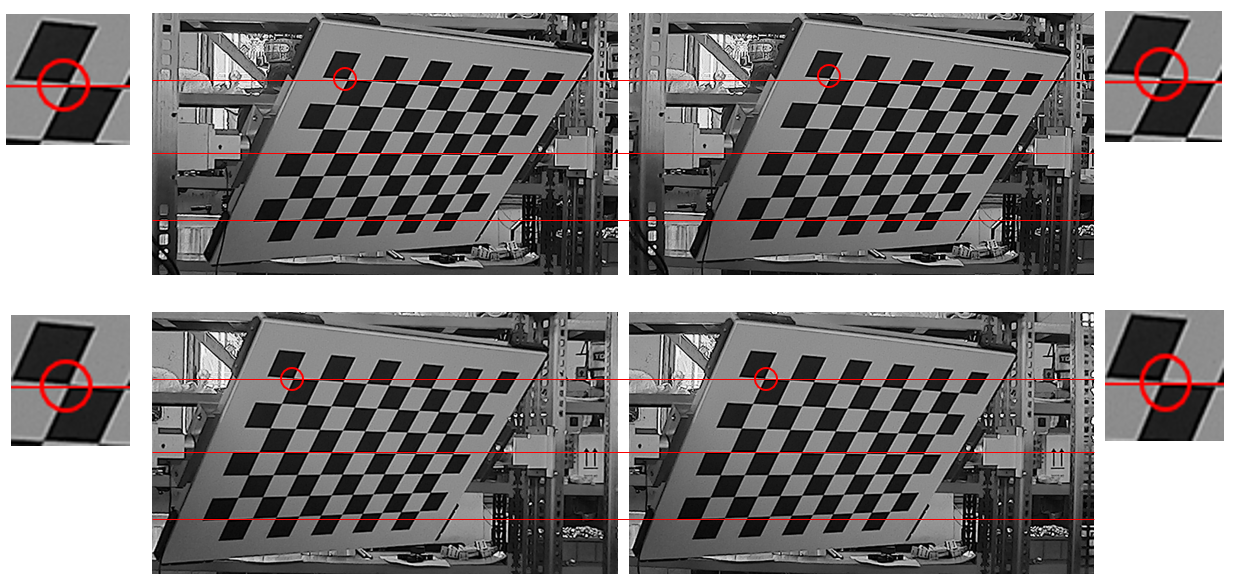}
	\caption{Stereo calibration.}
	\label{calibration_result}
\end{figure}

Image correction includes both the monocular distortion removing and binocular polar alignment. As shown in Figure \ref{calibration_result}, the top images (a) are origin images, in which the binocular polar is not aligned. We can see that the same corner points in the circles are not on the same line between the left and right images. The bottom images (b) are corrected images, in which the binocular polar has been aligned. We can see that the same corner points in circles are at the same line between the left and right images.

The undistorted image obtained by stereo calibration provides a good foundation for subsequent processing. Unless otherwise specified, the following images in this paper are all corrected images.

\subsection{Stereo Matching}
\begin{figure}[!h]
	\centering
	\includegraphics[width=14cm]{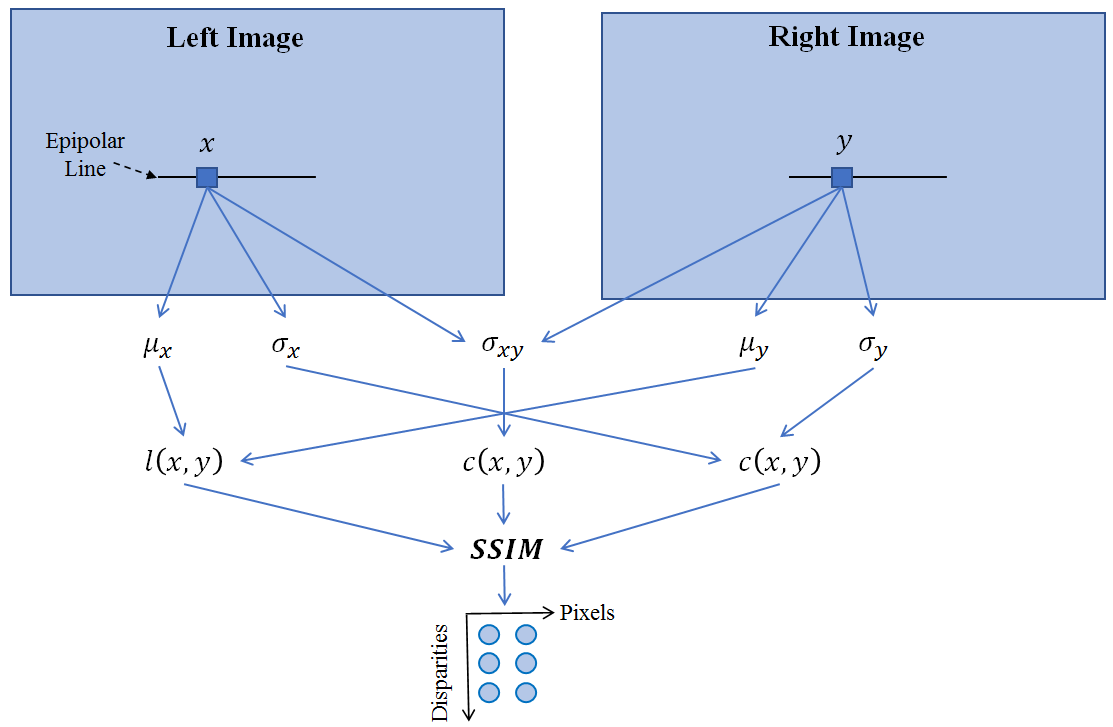}
	\caption{Fast calculation method of SSIM.}
	\label{fast_ssim}
\end{figure}

In this paper, after the calibration, all images conform to the pinhole image model and the coordinate origin is at the center of the image. We define that $I_0$ and $I_1$ are the rectified left and right images, and $\psi$ and $\phi$ are the $N \times N$ image patches located at $I_0$ and $I_1$, respectively. $\mu_\psi$, $\sigma_\psi^2$ and $\sigma_{\psi\phi}$ denote the mean, variance and covariance. Approximately, $\mu_\phi$ and $\sigma_\phi^2$ can be viewed as estimation of the luminance and contrast, and $\sigma_{\psi\phi}$ measures the tendency. This paper follows \cite{wang2004image}, and the luminance, contrast and structure similarity measures are given as follows:
\begin{equation}
	\begin{aligned}
	l(\psi,\phi) &= \frac{2\mu_\psi\mu_\phi  +C_1}{\mu^2_\psi+\mu^2_\phi + C_1} \\
	c(\psi,\phi) &= \frac{2\sigma_\psi\sigma_\phi + C_2}{\mu^2_\psi\mu^2_\phi + C_2} \\
	s(\psi,\phi) &= \frac{\sigma_\psi\sigma_\phi + C_3}{\sigma_{\psi\phi} + C_3}
	\end{aligned}
	\label{components}
\end{equation}
where $C_1$, $C_2$ and $C_3$ are constants given by $C_1=(K_1L)^2$, $C_2=(K_2L)^2$ and $C_3=C_2/2$, respectively. The $L$ is the dynamic range of the pixel values, where $K_1\ll1$ and $K_2\ll1$ are two scalar constants. The $SSIM$ cost function is defined as follow:
\begin{equation}
	SSIM(p,u) = \frac{(1-l^\alpha(\psi, \phi)c^\beta(\psi, \phi)s^\gamma(\psi, \phi))L}{2}
	\label{ssim}
\end{equation}
where $\alpha$, $\beta$ and $\gamma$ are parameters to define the relative importance of the corresponding three components. the $u$ is disparity and $p$ is pixel coordinates in image. We adopt a fast calculation method for $SSIM$, as shown in Figure \ref{fast_ssim}.

The MPV algorithm includes two parts. The first part estimates disparity by a Viterbi process and the second part is the path-merging strategy, which uses $4$ bi-directions (horizontal, vertical, and two diagonal) Viterbi paths on the matching space to provide good coverage of the $2D$ image.

We introduce total variation (TV) constraint in Viterbi path to constrain the disparity variation. Because TV constraint is applied to all the $4$ paths independently, $3D$ planes at different orientations can be approximately modeled by at least one path. Therefore, it can model the 3D objects with one or multiple slanted planes. TV constraint is useful to smooth some non-textured areas, such as roads or car bodies which are common in driving scenes but hard for stereo matching algorithms. Besides that, we also use the intensity gradient information to control the regularization level of TV constraint and make edges to be sharper.

\begin{figure}[!h]
	\centering
	\includegraphics[width=10cm]{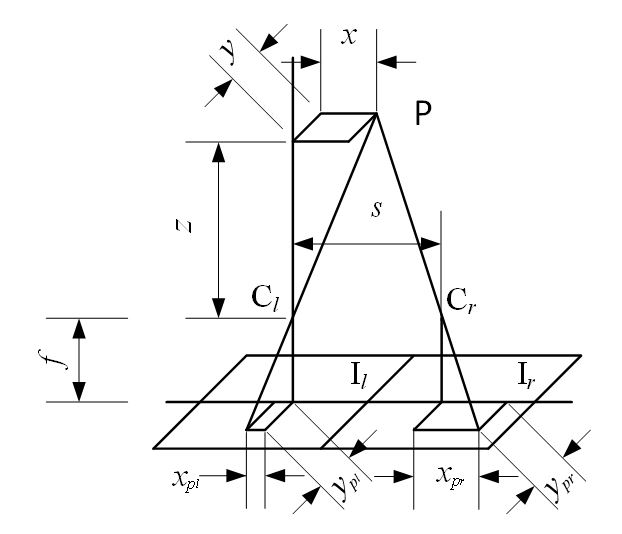}
	\caption{Total Variation (TV) constraint.}
	\label{tv}
\end{figure}

Figure \ref{tv} shows the principle of the TV constraint in our method. Both $C_l$ and $C_r$ are the focus of the left and right camera and $s$ is the baseline. The $P$ is an object point at a plane and the coordinate of $P$ satisfies the following formula:
\begin{equation}
	ax+by+cz=1
	\label{P}
\end{equation}

Denote $u$ as the disparity of $P$, then
\begin{equation}
    \begin{aligned}
	u &= |x_{pr} - x_{pl}| = s \cdot \frac{f}{z} \\
    x &= s \cdot \frac{x_{pl}}{u} \\
    y &= s \cdot \frac{y_{pl}}{u}	
    \end{aligned}
    \label{uxy}
\end{equation}

According to above hypothesis, we can obtain the following formula:
\begin{equation}
    \begin{aligned}
    a\cdot s\cdot\frac{x_{pl}}{u}+b\cdot s\cdot\frac{x_{pr}}{u}+c\cdot s\cdot\frac{f}{u} &= 1 \\
    \frac{u}{c\cdot s\cdot f}-\frac{a\cdot x_{pl}}{c\cdot f}-\frac{b\cdot x_{pr}}{c\cdot f} &= 1 \\
    \frac{\partial u}{\partial x_{pl}}=a \cdot s \qquad \frac{\partial u}{\partial x_{pr}}=b\cdot s
    \end{aligned}
    \label{5}
\end{equation}

For example, for a plane perpendicular to the optical axis $z=z_0$ , then $a=0$ and $b=0$. In addition, we take the ground plane as an example $y=y_0$, then $a=0$. Since ADAS focuses on the road surface and the corresponding obstacles, it is reasonable to add TV constraint, which is expressed by defining the energy $E(U)$ on the disparity map $U$ as follows:
\begin{equation}
    \begin{aligned}
    E\left(U\right) &= \sum_{p} S S I M\left(p,u\right)+\sum_{p^\prime\in L_p} s_{\left(p^\prime,u^\prime\right)\rightarrow\left(p,u\right)} \\
    s_{\left(p^\prime,u^\prime\right)\rightarrow\left(p,u\right)} &= \lambda e^{-\left|G\right|}\left|u-u^\prime\right|\
    \end{aligned}
    \label{EU}
\end{equation}
where $s$ is the TV constraint modified by the gradient $G$ of image $I_0$. It penalizes all the disparity changes between $p$ and $p'$, where $p'$ has disparity $u'$ and belongs to the neighborhood $L_p$ of $p'$. The $\lambda$ is the tradeoff parameters to balance the TV term and the fit term.

\begin{figure}[!h]
	\centering
	\includegraphics[width=14cm]{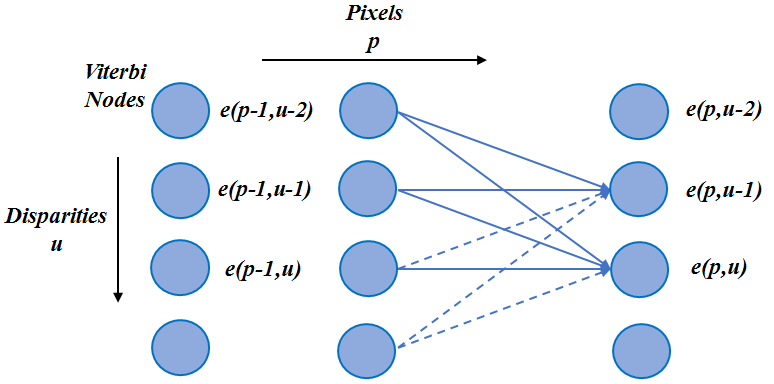}
	\caption{Trellis diagram for nodes and edges in a same Viterbi path.}
	\label{nodes_edge}
\end{figure}

The stereo matching solution can be formulated as finding the disparity map $U$ that minimizes the energy function $E(U)$. The Viterbi algorithm can be used to approximate the optimum solution \cite{son2006stereo}. In this case, the Viterbi trellis represents a graph of disparity states for all pixels. Each node in this trellis represents a disparity assigned to a pixel and each edge represents a possible disparity change between two adjacent pixels in the same Viterbi path, as shown in Figure \ref{nodes_edge}. We define the energy of a node as $e(p, u)$ with pixel $p$ and disparity $u$, as following.
\begin{equation}
    \hat{U} = \underset{U}{arg\min}E(U) \approx \underset{U}{arg\min} \underset{all\ Viterbi\ path}{\sum}e(p,u)
    \label{viterbi}
\end{equation}

According to Viterbi algorithm:
\begin{equation}
    e(p,u) = \underset{u'\in L_u}{min}\left\lbrace e(p-1,u')+\varepsilon_{(p-1,u')\in (p,u)}+SSIM(p,u) \right\rbrace
     \label{e(p,u)}
\end{equation}
where, $L_u$ means the connected nodes from $(p-1, u?．)$ to $(p, u)$, and $(p-1)$ means the previous node at the same Viterbi path. In normal Viterbi algorithm, node number of $L_u$ is generally small and the total computational cost for one pixel is $O(N(u) \times N(Lu))$, where $N(\cdot)$ indicates the number of nodes. In our MPV algorithm, we set the $L_u$ as all the possible Viterbi nodes. This setup can keep edge sharp for outdoor scenes but it increases the computational cost to $O(N(u)^2)$.

\begin{figure}[!h]
	\centering
	\includegraphics[width=14cm]{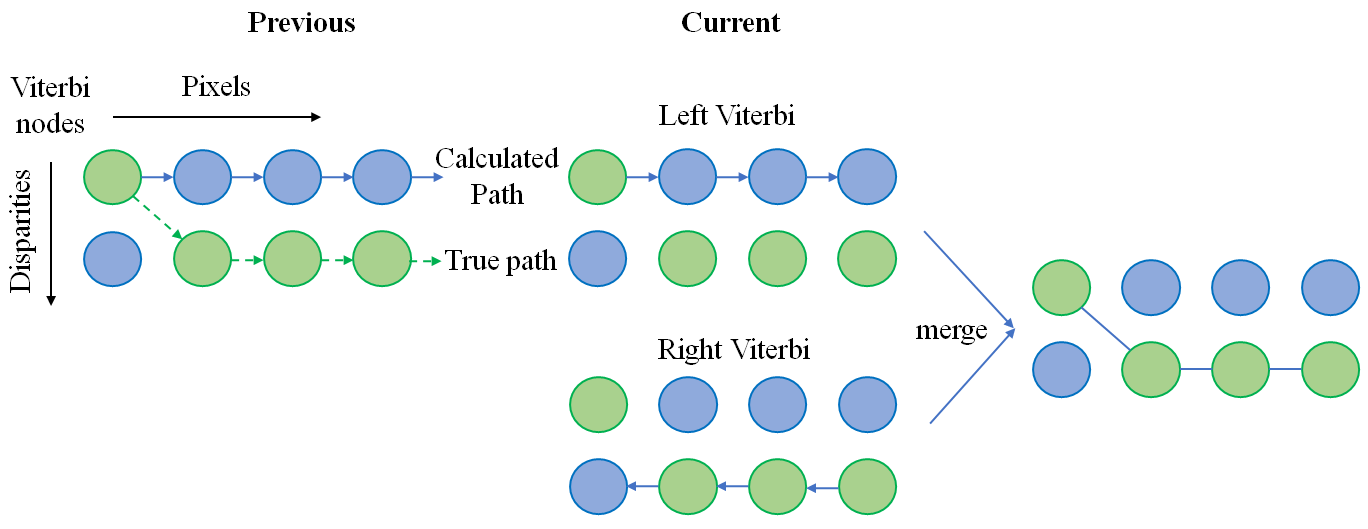}
	\caption{Merging strategy in Viterbi.}
	\label{merging}
\end{figure}

In each layer, we apply bi-directional Viterbi algorithm according to Equation \ref{5}. Then, we update the Viterbi nodes energy by using optimum energy of the two opposite directions. For horizontal path, we use the minimum function to sharpen edges,  and  for  other  paths  we  use  the  average  function  to remove noises. After finishing one layer, the energy of Viterbi nodes of current layer is used as the initial value of the energy of Viterbi nodes at the next layer. Here, several specific strategies can be applied to the path merging for every layer. As shown in Figure \ref{merging}, for example, we set a twice penalty to the left Viterbi in case of changing from small disparity to big one, which helps to improve the performance at occluded area.

\begin{figure}[!h]
	\centering
	\includegraphics[width=14cm]{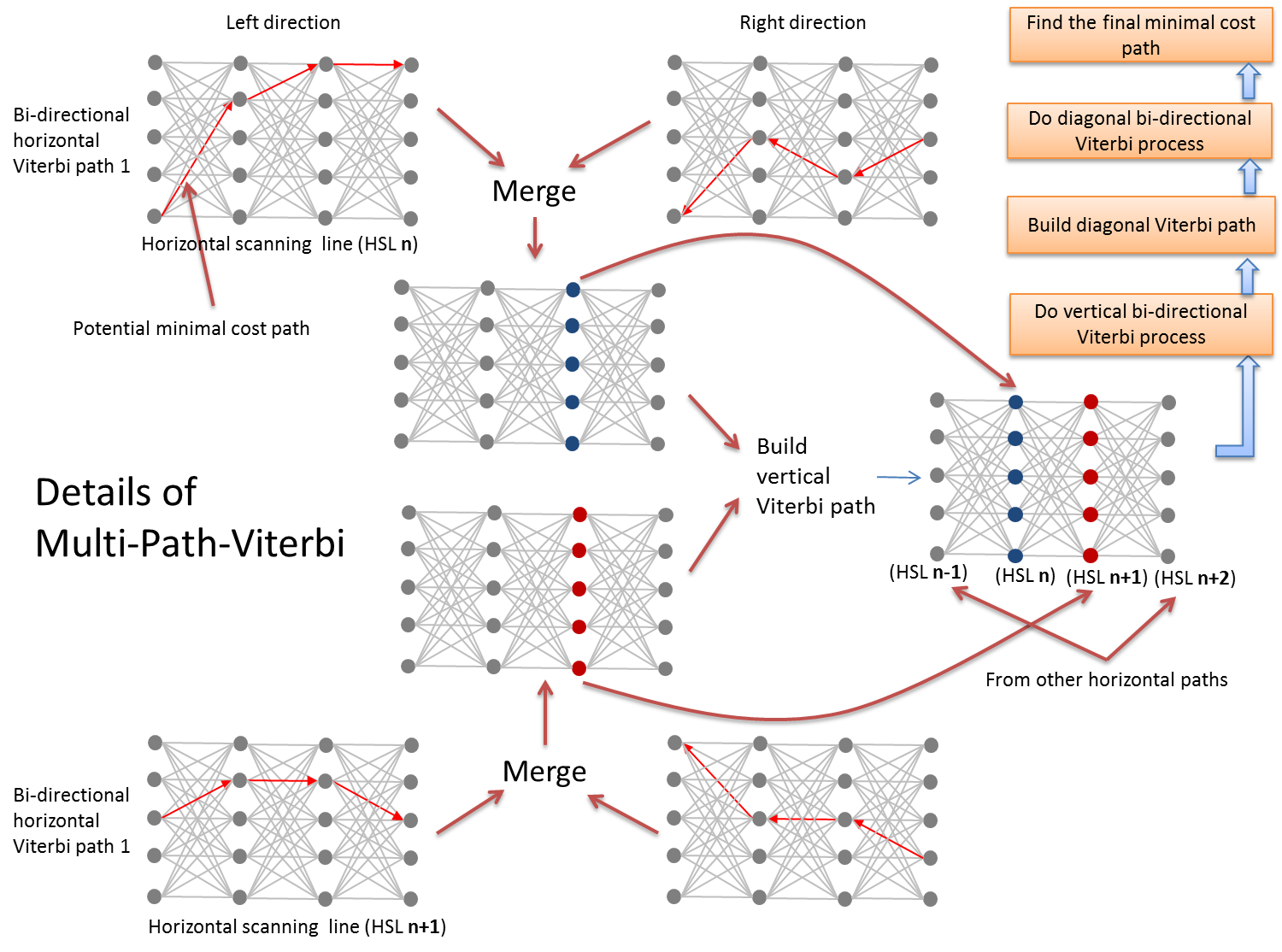}
	\caption{Hierarchical structure for the merging of multiple Viterbi paths.}
	\label{viterbi_path}
\end{figure}

\begin{figure}[!h]
	\centering
	\includegraphics[width=14cm]{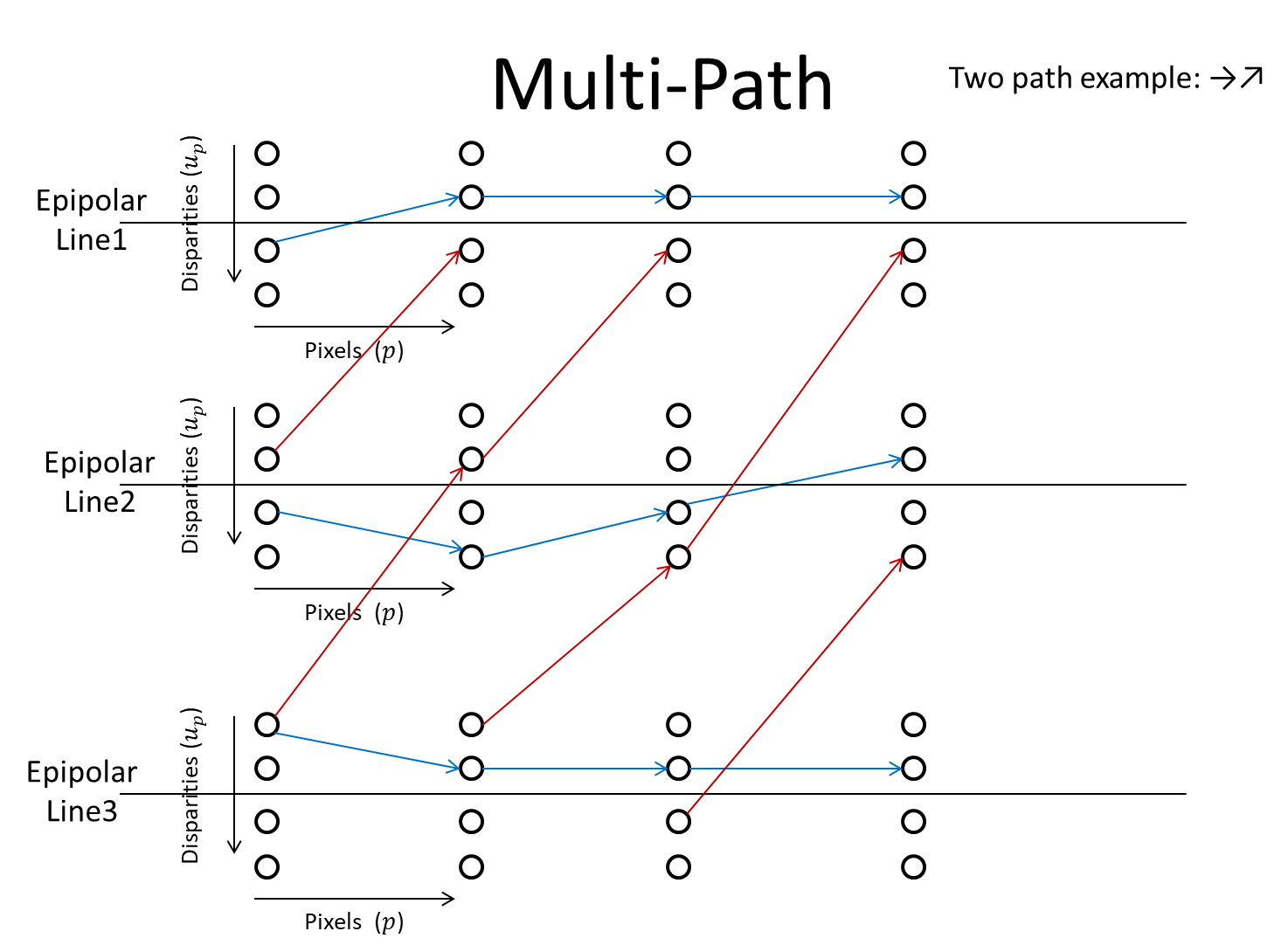}
	\caption{Hierarchical structure for the merging of multiple Viterbi paths.}
	\label{search_path}
\end{figure}

As shown in Figure \ref{viterbi_path}, we use $4$ bi-directional (horizontal, vertical, and two diagonals) Viterbi paths on the matching space to provide good coverage of the $2D$ image. Horizontal directions have stronger constraints compared to other directions. In our approach, we use the results of horizontal directions as strong posterior information to calculate the optimum paths of other directions. There are $4$ hierarchical layers in this paper, and we apply bi-directional Viterbi algorithm in each layer. Then, we update the Viterbi node?．s energy by using optimum energy of the two opposite directions. We refine the Multi-Path-Viterbi algorithm as Table \ref{alg} and Figure \ref{search_path} shows the example for search paths.

\begin{table}
	\centering	
	\caption{Multi-path viterbi algorithm.}
	\begin{tabular}[!h]{l}		
		\toprule
		\textbf{Algorithm:} Multi-path viterbi algorithm \\
		\midrule
		\textbf{Input:}  Previous Pixels\\
        \textbf{Output:}  new Viterbi energy\\
        Step 1: Compute the left and right bi-directional Viterbi algorithm;\\
        Step 2: Compute the up and down bi-directional Viterbi algorithm;\\
        Step 3: Compute the right down and left up Viterbi algorithm;\\
        Step 4: Compute the left down and right up Viterbi algorithm.\\
		\bottomrule	
	\end{tabular}
    \label{alg}
\end{table}

\subsection{Fast calculation technique to find best viterbi path}
\begin{figure}[!h]
	\centering
	\includegraphics[width=10cm]{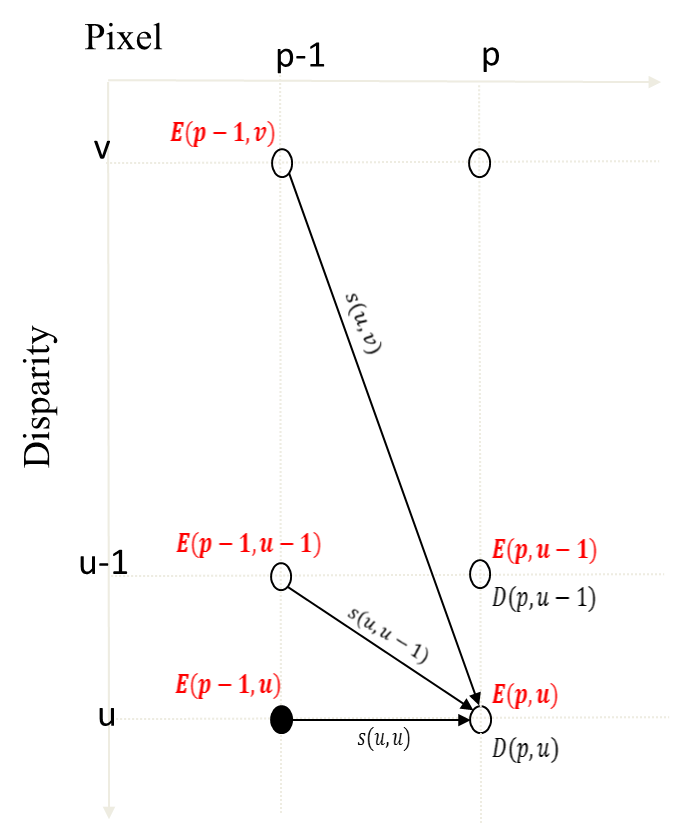}
	\caption{Simplified technique for the Viterbi searching path.}
	\label{simplified_tech}
\end{figure}

In order to implement the algorithm on real-time system, we propose a new fast calculation technique to find the best Viterbi path. For normal Viterbi algorithm on epipolar line, if searching n branches for m disparity nodes, it needs $O(nm)$ searching. Now we can search m branches for m disparity nodes for $O(2m)$ searching. This new fast calculation technique is based on a new simplified searching path, as shown in Figure \ref{simplified_tech}, which is derived from the following theorem.

The original Viterbi searching form (n term for comparison in minimum function) is Eq.\eqref{eq1}:
\begin{equation} \label{eq1}
    E(p,u) = \min_{v \in L_u}\{E(p-1,v)+\varepsilon(u,v)+SSIM(p,u)\}
\end{equation}

We proposed fast calculation alternative formula as Eq.\eqref{eq2}:
\begin{equation} \label{eq2}
\begin{aligned}
    E(p,u) = \min_{v \in L_u}\{&E(p,u-1)-SSIM(p-1,u)-\varepsilon(u-1,v)+\varepsilon(u,v), \\
    & E(p-1,u)+\varepsilon(u,v)+SSIM(p.u)\}
\end{aligned}
\end{equation}

If replace $u$ as $u-1$, we obtain Eq.\eqref{eq3}:
\begin{equation} \label{eq3}
	E(p,u-1)=\min_{v \in L_u}\{E(p-1,u)+\varepsilon(u-1,v)+SSIM(p,u-1)\}
\end{equation}

We can get as Eq.\eqref{eq4}:
\begin{equation} \label{eq4}
	E(p,u-1)-SSIM(p,u-1)=\min_{v \in L_u}\{E(p-1,v)+\varepsilon(u-1,v)\}
\end{equation}

If we add $1$ on both sides of the equation as Eq.\eqref{eq5}:
\begin{equation} \label{eq5}
\begin{aligned}
	& E(p,u-1)-SSIM(p,u-1)+1 \\
	&= \min_{v \in L_u}\{E(p-1,v)+s(u-1,v)\}+1 \\
	&= \min_{v \in L_u}\{E(p-1,v)+s(u-1,v)+1\}
\end{aligned}
\end{equation}

Based on the defintion $\varepsilon(u,v)$, we derive the equation as Eq.\eqref{eq6}:
\begin{equation} \label{eq6}
	\varepsilon(u,v) = 1 + \varepsilon(u-1,v)
\end{equation}

Combined Eq.\eqref{eq5} and Eq.\eqref{eq6}, we can obtain Eq.\eqref{eq7}:
\begin{equation} \label{eq7}
\begin{aligned}
    E(p,u-1)-SSIM(p,u-1)-s(u-1,v)+\varepsilon(u,v)  \\
    = \min_{v \in L_u}\{E(p-1,v)+\varepsilon(u-1,v)-\varepsilon(u-1,v)+\varepsilon(u,v)\}
\end{aligned}
\end{equation}

Integrated Eq.\eqref{eq7} with Eq.\eqref{eq3}, we get Eq.\eqref{eq8}:
\begin{equation} \label{eq8}
\begin{aligned}
    E(u,p) = \min_{v \in L_u}\{&E(p,u-1)-SSIM(p-1,u)-\varepsilon(u-1,v)+\varepsilon(u,v), \\
     &E(p-1,u)+\varepsilon(u,v)+SSIM(p.u)\}
\end{aligned}
\end{equation}

Based on the above theorem, we reduce the complexity from $O(nm)$ to $O(2m)$ for normal Viterbi algorithm on epipolar.

\subsection{Multi-scale image matching approach}
We refer to the multi-scale image matching approach as another fast calculation technique. Inspired by \cite{lowe2004distinctive,date2019level}, two pieces of main recommendations are made: image pyramid and multi-scale disparity transformation. We reduce the image size by implementing down-sampling and the disparity from the previous layer is passed and transformed to next layer. We only calculate the matching value of full range for each pixel at the top of the pyramid, while more pixels at other layers are obtained by disparity transformation. The specific principle is as follows.

Firstly, down-sample the image to a preset scale. In this paper, we set the preset scale to three layers. That is, the image at the top of the pyramid (Layer $0$) is one-sixteenth of the original image (Layer $2$, the middle layer is Layer $1$). Therefore, the matching range in the top layer is one quarter of the matching range of the original image.

Next, we split the image at the $0^{th}$ layer to some blocks which have the same pixels size. We implement the MPV algorithm for each block. As a result, every block has the initial disparity value. We suggest that the size of these block at the $0^{th}$ layer should not be too small, because there are mainly large feature objects.

Then, we pass these initial disparity values to the next layer. Obviously, not all pixels have initial disparity values at the $1^{st}$ layer, and only the pixels sampled to the previous layer have initial values. We continue to split images to smaller blocks than the previous layer. We suggest that the closer the image layer is to the original image, the smaller the size of the block, otherwise, more pyramid layers will need to be created. The mode of initial disparity at each small block is assigned as the initial value of the MPV algorithm. It should be noted that these initial values have to be transformed by the sampling rate.

Besides, the searching scope of the MPV algorithm is dynamically adjusted in different layer. Therefore, the searching scope of each layer is inconsistent. For current layer, we add some virtual nodes to support the MPV algorithm.

At last, when disparity value is transformed to the last layer, original images, we split each pixel as a block and initialize the uninitialized pixels by linear interpolation .
Assume that the size of the image is $m \times n$ and the maximum searching scope is $d$ pixels, the maximum algorithm complexity of the original MPV algorithm is:
\begin{equation}
O(t)=m\times n\times d = mnd.
\label{Ot}
\end{equation}

For the $3$-scale MPV algorithm, the complexity of the $0^th$ layer is:
\begin{equation}
O\left(t_0\right)=\frac{m}{4}\times\frac{n}{4}\times\frac{d}{4} = 0.0156mnd.
\label{Ot0}
\end{equation}

The complexity of the $1^{st}$ layer is:
\begin{equation}
O\left(t_1\right)=\frac{m}{2}\times\frac{n}{2}\times\frac{d}{4}=0.0625mnd.
\label{Ot1}
\end{equation}

And the complexity of the $2^{nd}$ layer is:
\begin{equation}
O\left(t_2\right)=m\times n\times\frac{d}{4} = 0.25mnd.
\label{Ot2}
\end{equation}

Therefore, the total maximum algorithm complexity of the $3$-scale MPV algorithm is:
\begin{equation}
O\left(t\right)=\ O\left(t_0\right)+O\left(t_1\right)+O\left(t_2\right)=0.3281mnd
\label{O}
\end{equation}

Comparing Eq. \eqref{Ot} and Eq. \eqref{O}, it is obvious that our improved algorithm is superior to the original one.

\begin{figure}[!h]
	\centering
	\includegraphics[width=12cm]{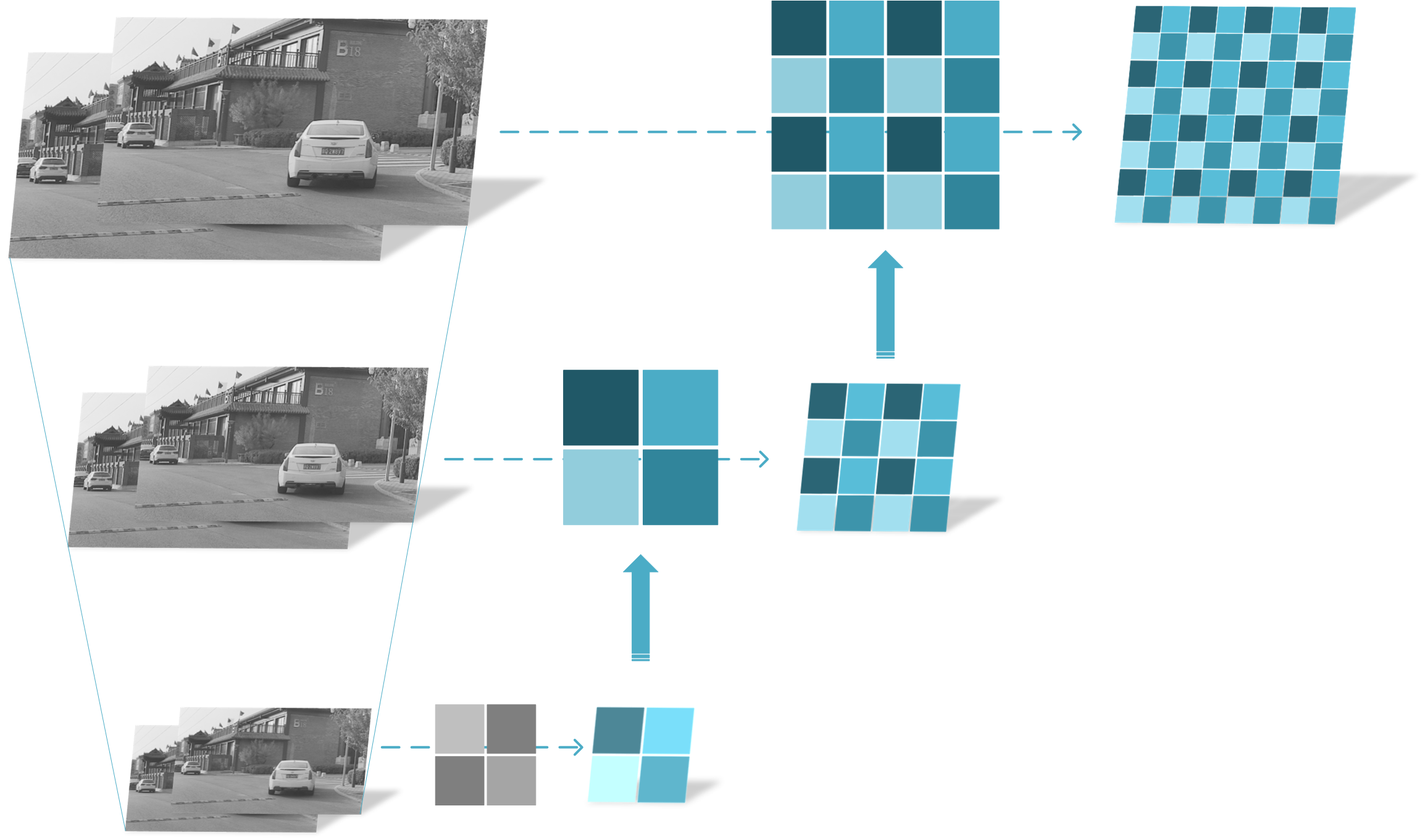}
	\caption{Multi-scale technique for the MPV algorithm.}
	\label{multi_scale}
\end{figure}

\begin{figure}[!h]
	\centering
	\includegraphics[width=14cm]{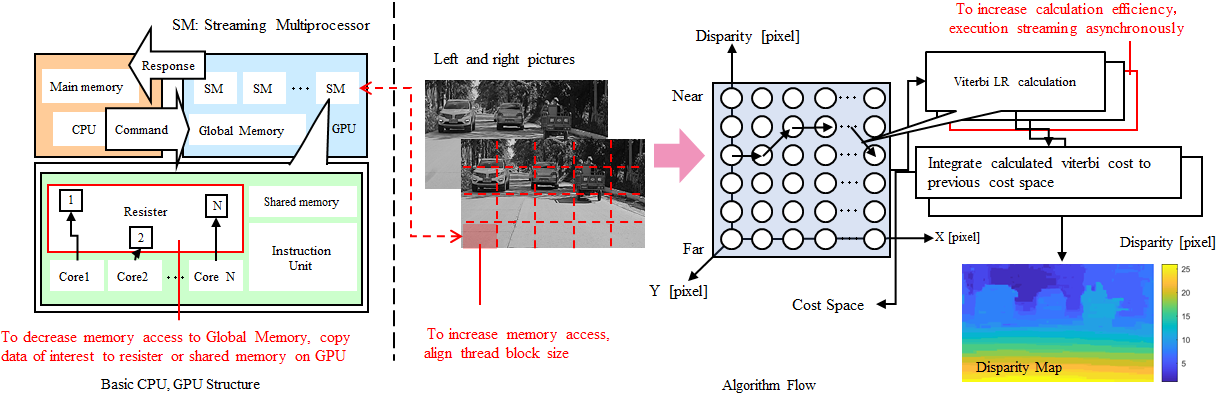}
	\caption{Toward Real-time Disparity Estimation.}
	\label{disp_estimation}
\end{figure}

Figure \ref{multi_scale} shows the proposed multi-scale technique for the MPV algorithm. The bottom of Figure \ref{multi_scale} demonstrates the implementation of the multi-scale technique. The four previous blocks have the same searching scope and different disparities (different colors). Each large block is split into four small blocks on current layer with the different initial disparities, and these small blocks have different searching scope (different colors). As the contrast, the small blocks have the same search scope (same color) in the original algorithm shown in upper right corner. Figure \ref{disp_estimation} illustrates our implementation platform of above fast calculation technique.

\subsection{Road and obstacle detection}
Using the disparity map $U$ of the image, we can calculate the histogram of disparity map in horizontal and vertical directions. Then $H_x\{U\}(i,j)$ corresponds to the number of points with same disparity as $i$ at the horizontal image line $j$ in the disparity map $U$:
\begin{equation}
H_x\left\{U\right\}\left(i,j\right) = \sum_{\left(x,y\right)\in U}{\delta_{y,j}\delta_{u\left(x,y\right),i}}
\label{disp_map}
\end{equation}
where $\delta$ denotes the Kronecker delta. Similarly, for any pixel $(i,j)$ in $H_y{U}$, we have
\begin{equation}
H_y\left\{U\right\}\left(i,j\right) = \sum_{\left(x,y\right)\in U}{\delta_{x,j}\delta_{u\left(x,y\right),j}}
\label{hy}
\end{equation}

Generally, we can detect the road model in $H_x\{U\}$ and detect the obstacle in $H_y\{U\}$. Radon transform is performed to detect road. Given disparity map $U$, Randon transform to $R\{U\}$ can be defined as:
\begin{equation}
R\left\{U\right\}\left(d,\phi\right)=\int_{R^2}{H_x\left\{U\right\}\left(i,j\right)\varepsilon\left(icos{\phi}+jsin{\phi}-d\right)didj}
\label{Ru}
\end{equation}
where $\varepsilon$ denotes Dirac delta function, $d$ is the distance from the line to the origin through the normal of the line that intersects the origin, and $\phi$ is the angle between the same normal and x-axis. $R\{U\}(d, \phi)$ provides a mapping from $H_x\left\{U\right\}$ to a parameter space spanned by $d$ and $\phi$.

Viterbi searching space is built in the $H\{U\}$ directly to detect the road and curb. We treat every pixel $(i, j)$ in the $H\{U\}$ as a Viterbi node and the Viterbi process accumulates the value of each node to find a continuous path $j=P(i)$ that has the maximum sum of value at proper straightness constraints. For road detection, the Viterbi equation is:
\begin{equation}
\hat{P} = argmax{e(i,P(i))}
\label{P_hat}
\end{equation}

According to Viterbi algorithm, we have:
\begin{equation}
e(i,j) = H_x\{U\}(i,j)+\max_{0<j-j^\prime<\eta} e(i-1,j^\prime)
\label{e}
\end{equation}
where $\eta$ is the parameter to control the straightness of road.

With the road area G and disparity map $U$, we can map every point in the G to $3D$ space with the camera parameters. Let $(x, y)$ denotes image coordinate of a pixel in the image and u denotes its disparity value such as that $u = U(x, y)$. Assume its coordinate is $(X, Y, Z)$. Given the focal length $f$ and base line length $B $of calibrated stereo vision system, we have $u/B = f/Z = x/X = y/Y$ according to the geometry of stereo vision.

Suppose the height of the small object is $S_h$ and the equation of the road surface is $ax + by + cz = d$. Then we can classify a pixel $(x, y)$ as a part of a small object on road if and only if:
\begin{equation}
(x,y)\in G\land0 < \frac{axB+byB+cfB-du}{\sqrt{a^2u^2+b^2u^2+c^2u^2}}<S_h
\label{sh}
\end{equation}

\subsection{Target recognition}
After road and obstacle being detected in the disparity map, we can obtain the ROI windows \cite{choi2012environment,caraffi2007off,altun2017road} of the obstacles. Then, we extract the area where the window located in the left image \cite{mammeri2016extending,yuan2015multisensor,lins2015vision}. In this case, obstacles are provided to the target recognition system.

\begin{figure}[!h]
	\centering
	\includegraphics[width=12cm]{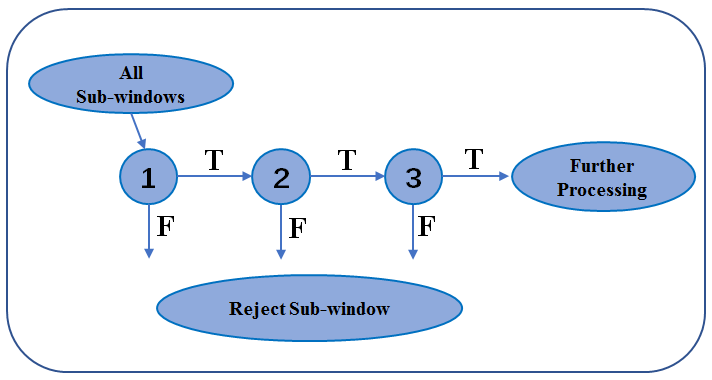}
	\caption{Schematic depiction of a detection cascade.}
	\label{cascade}
\end{figure}

The task of monocular target recognition is released by cascade classifiers based on the AdaBoost algorithm \cite{schapire2013explaining}. In general, most weak classifiers can be used to construct a cascade. The key properties are that the computation time and detection rate can be adjusted \cite{liao2015fast}. We train each weak classifier by AdaBoost until the detection accuracy meets certain constraints \cite{li2008adaboost}. The process of cascade \cite{viola2002fast} is shown in Figure \ref{cascade}. The initial classifier eliminates many negative examples with very little processing. Subsequent stages eliminate additional negatives but require additional computation. After several stages, few extreme negative examples remain.

We train the cascade classifier with the LBP feature \cite{wang2009hog}. This feature is similar to a texture descriptor for object detection. According to \cite{liu2016median}, we set a weight of $2n$ to the central pixel. In the \cite{aguilar2014robust}, the parameters of LBP operator are $p$ and $r$, where $p$ is the number of pixels and $r$ is the distance to the central pixel. Then the LBP is defined as follow:
\begin{equation}
LBP_{p,r}\left(X_c,Y_c\right) = \sum_{p=0}^{p-1}{s\left(g_p-g_c\right)\cdot2^p}
\label{lbp}
\end{equation}
where $g_p$ is the value of center pixel, $g_c$ is the value of neighbor, and $2^p$ is the weight for each operation. The $s(g_p-g_c)$ can be defined as follows:
\begin{equation}
s(g_p-g_c) = \left\{\begin{matrix}1&g_p-g_c\geq0\\0&g_p-g_c<0\\\end{matrix}\right.
\label{s}
\end{equation}

Finally, we structure a cascade, shown as follows, of a linear combination with selected weak classifiers.
\begin{equation}
C\left(x\right)=\sum_{t=1}^{T}{\xi^t h^t(x)}
\label{C}
\end{equation}
where, $h^t(x)$ is a weak classifier and $\xi^t$ is a weight coefficient. Based on above weak classifiers, we can construct a superior performance cascade AdaBoost classifier in real-time \cite{hu2013online}.

\subsection{Distributed computing technology}
On our platform, GPU has powerful graphics computing capabilities. The hardware calculation flow of the fast SSIM algorithm is shown in the Figure \ref{disp_estimation}, which can employ the GPU power and ensure that our system is a real-time processing system.

\begin{figure}[!h]
	\centering
	\includegraphics[width=12cm]{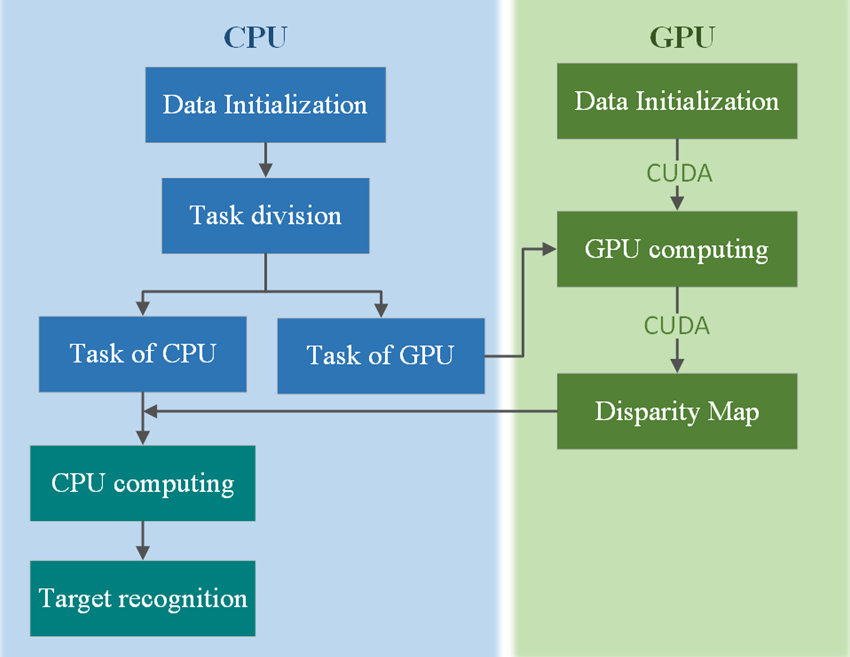}
	\caption{Task segmentation based on distributed computation.}
	\label{distributed_comptation}
\end{figure}

We divide the task of sensing the $3D$ environment into two parts that are disparity map calculation and target recognition. Distributed computation is implemented in our system to complete parallel computing of two parts at the same time. We calculate the disparity map by GPU and implement the target recognition by CPU. Distributed computation \cite{thain2005distributed} process is shown in Figure \ref{distributed_comptation}.

Binocular images are input to GPU where disparity map is calculated. Then, GPU feeds the disparity map and left image into ARM in which target recognition is implemented. At the same time, GPU is computing the next disparity map \cite{zhang2012imapreduce}. In addition to target recognition, ARM carries out system control for the rest of the time \cite{kshemkalyani2011distributed}. In this way, our system can achieve real-time detection.

\begin{figure}[!h]
	\centering
	\includegraphics[width=14cm]{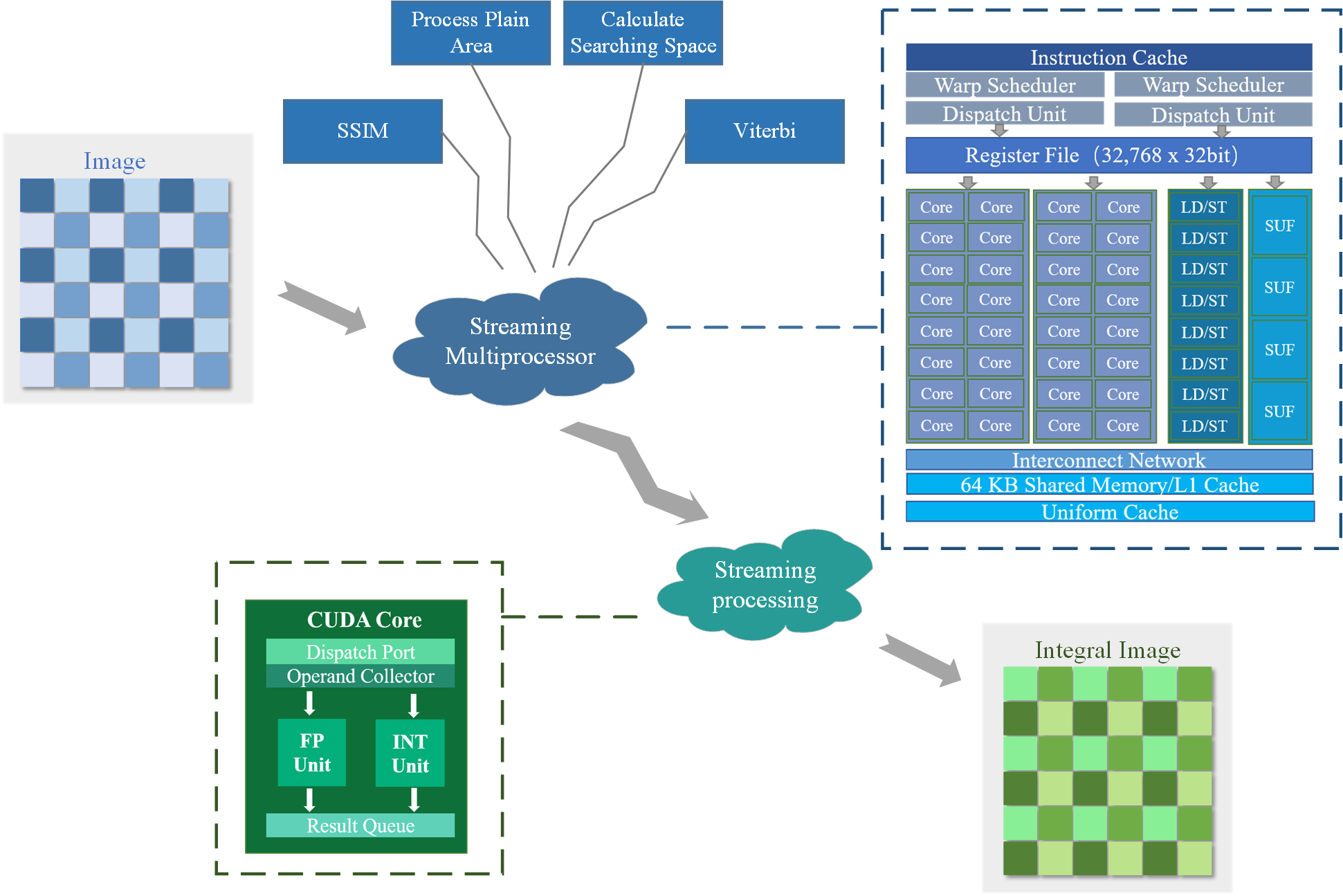}
	\caption{GPU acceleration flow.}
	\label{gpu_flow}
\end{figure}

In our method, a lot of complicated calculations are done by GPU. Accelerated by GPU, our stereo matching algorithm can be implemented less than $80$ms. GPU acceleration unit is shown in Figure \ref{gpu_flow}. The left flow chart is the process of GPU calculation, and the right frame diagram is the schematic diagram of GPU hardware acceleration.

\subsection{Data flow in system}
The detail of distributed computation can be divided into four steps as following.

\begin{itemize}
  \item Two images are stored in shared memory. The CPU sends instructions to GPU to evaluate disparity. The GPU fetches images from shared memory to implement disparity evaluation through internal distributed processing. Then, the disparity map is pushed into shared memory by GPU.

  \item The CPU catches the disparity map for road detection. Furthermore, the alternative region is extracted from left image and stored in shared memory.

  \item The ROI is assigned to other CPUs, where a machine learning model is carried out to obstacle detection.
\end{itemize}

\begin{figure}[!h]
	\centering
	\includegraphics[width=12cm]{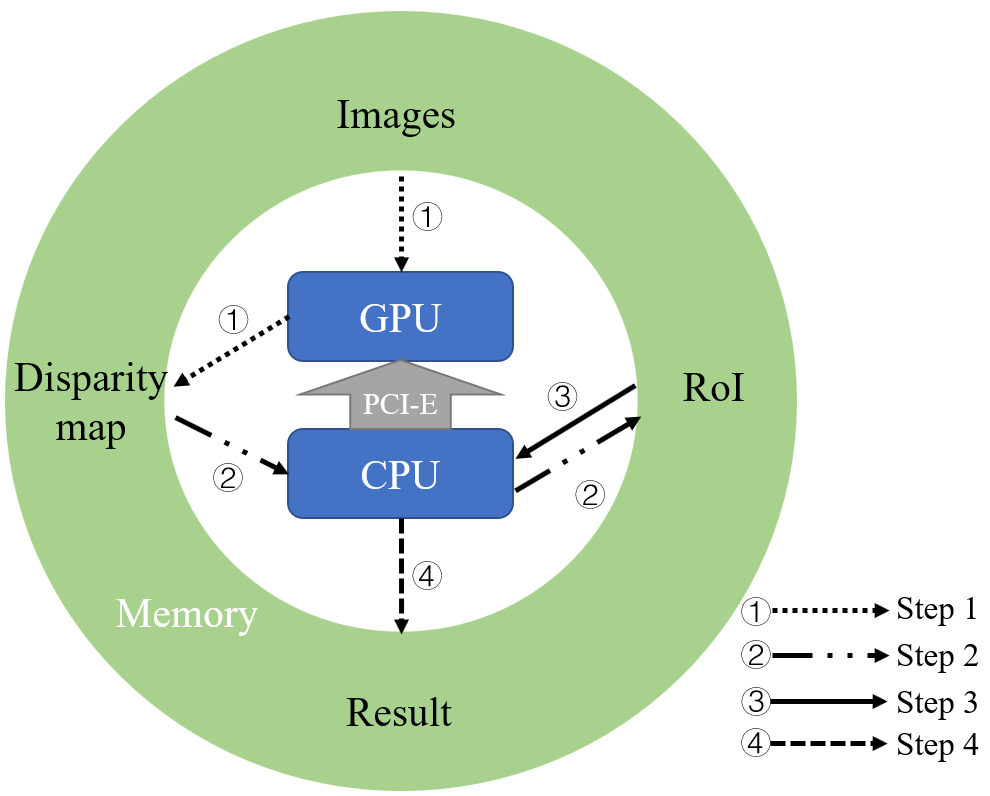}
	\caption{Data flow of distributed computation.}
	\label{data_flow}
\end{figure}

\begin{figure}[!h]
	\centering
	\includegraphics[width=12cm]{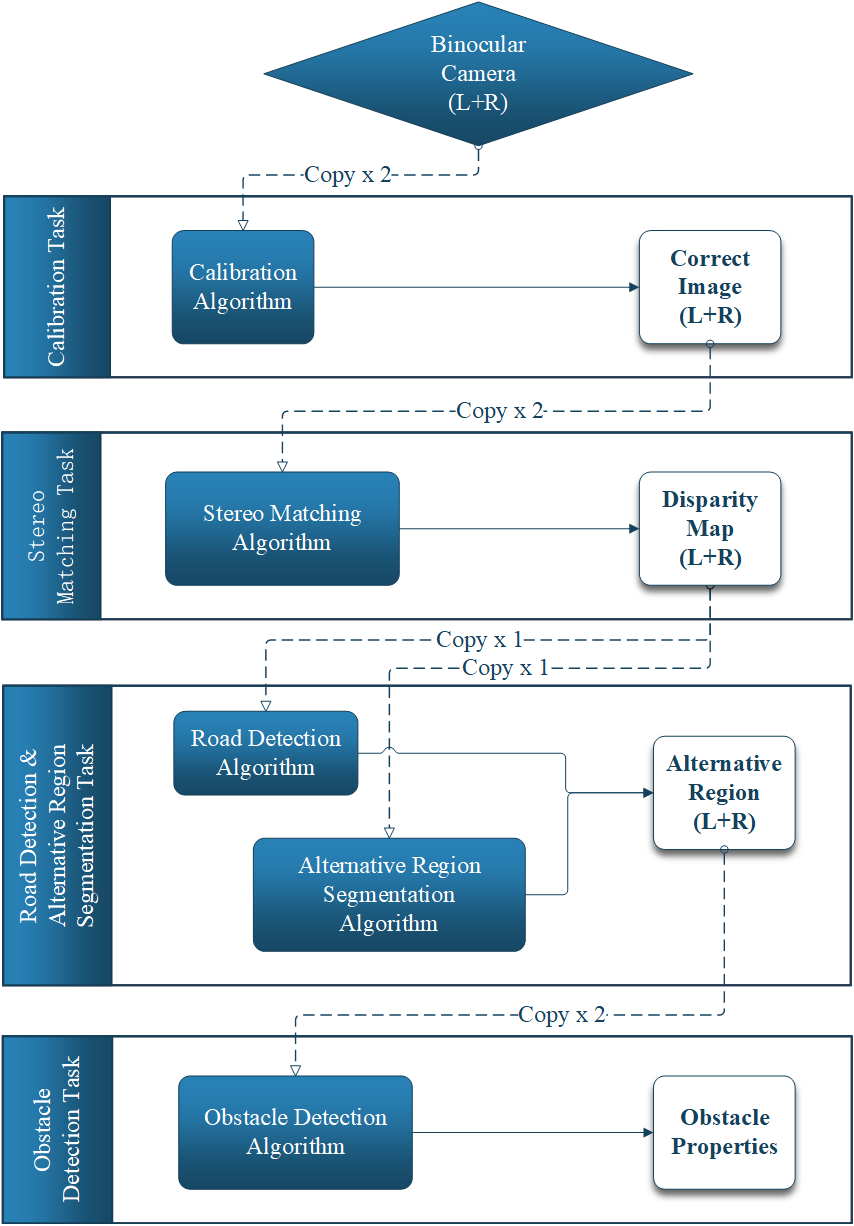}
	\caption{Data flow in task module.}
	\label{task_flow}
\end{figure}

As shown in Figure \ref{data_flow}. Since our system platform TK$1$ has $4$ CPUs and $1$ GPU, except for stereo matching which is calculated by GPU, other tasks are shared by $4$ CPUs. The data flow is shown in Figure \ref{task_flow}.

The frequency of GPU is dynamically adjusted, in order to save power cost, we set the frequency of GPU at a fixed value. The Fast-MPV algorithm is implemented under the fixed frequency by GPU.

Shared memory is used between processes and is managed by the system. Threads can use memory address passing to access the same buffer, but we copy a new memory, because of that when more than two threads access the same buffer at the same time, they are controlled by system lock that takes a certain time to apply to the system (more than $10$ms). Assume that the frame rate of image acquisition is $20$fps and the processing speed is $12.5$fps, the maximum number of memory copies in 1s is shown in Eq. \eqref{copies}:
\begin{equation}
2\times13+(2+2+2+2+1)\times20+4\times2=214
\label{copies}
\end{equation}

For a $640 \times 480$ image, the memory copy speed is less than $0.3$ms, which is determined by testing. Therefore, $214$ memory copies cost about less $20$ms and they distributed in $4$ CPUs.

\section{Experiment and Analysis}
In this section, we conduct four types of experiments to evaluate the performance of our system. First, we evaluate the performance of our stereo matching algorithm by using the KITTI dataset. Then the detection accuracy and running time, the entire system test, and the performance of the stereo matching algorithm under different weather and lighting conditions are evaluated by using our own collected real-time driving datasets. At last, the system hardware performance is tested.

\subsection{Evaluation of Stereo Matching Algorithm}
\begin{figure}[!h]
	\centering
	\includegraphics[width=16cm]{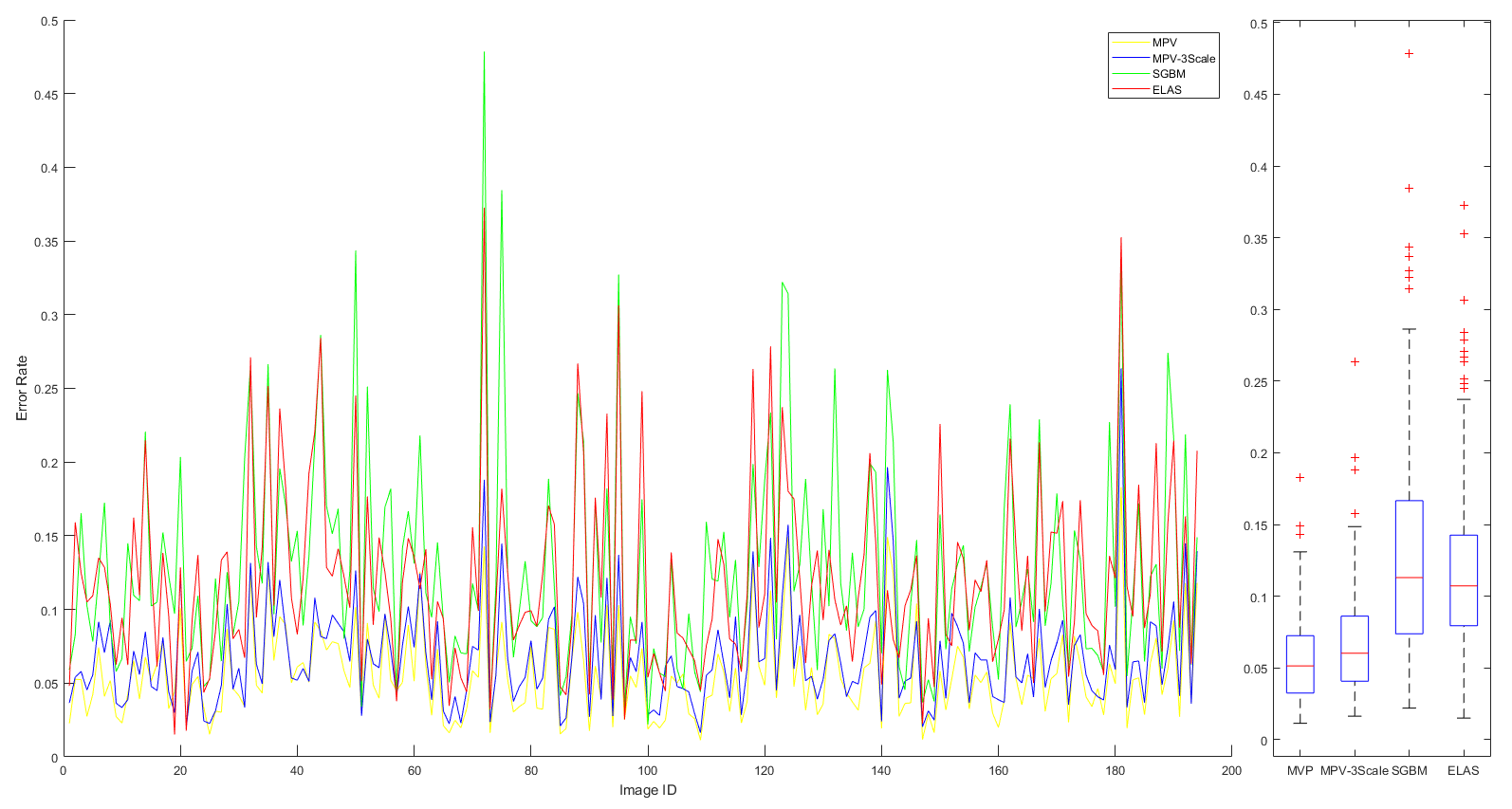}
	\caption{Error Rate of stereo matching algorithm.}
	\label{kitti}
\end{figure}

The performance of our proposed stereo matching algorithm is evaluated by using the KITTI datasets \cite{geiger2012we}. We use the KITTI training dataset which includes $194$ images and use the development kit in the KITTI website to do the evaluation. The MPV algorithm has $5.57\%$ average error rate and our fast-MPV algorithm has $6.78\%$ average error rate, which outperforms SGBM with $12.88\%$ \cite{hirschmuller2007stereo} and ELAS with $11.99\%$ \cite{geiger2010efficient}. These results are illustrated in Figure \ref{kitti}. Although the fast-MPV algorithm's error rate is slightly higher than the original MPV algorithm, it is still significantly better than the SGBM and ELAS. At the same time, the computation cost of the fast algorithm is far less than that of the original algorithm.

\begin{figure}[!h]
	\centering
	\includegraphics[width=12cm]{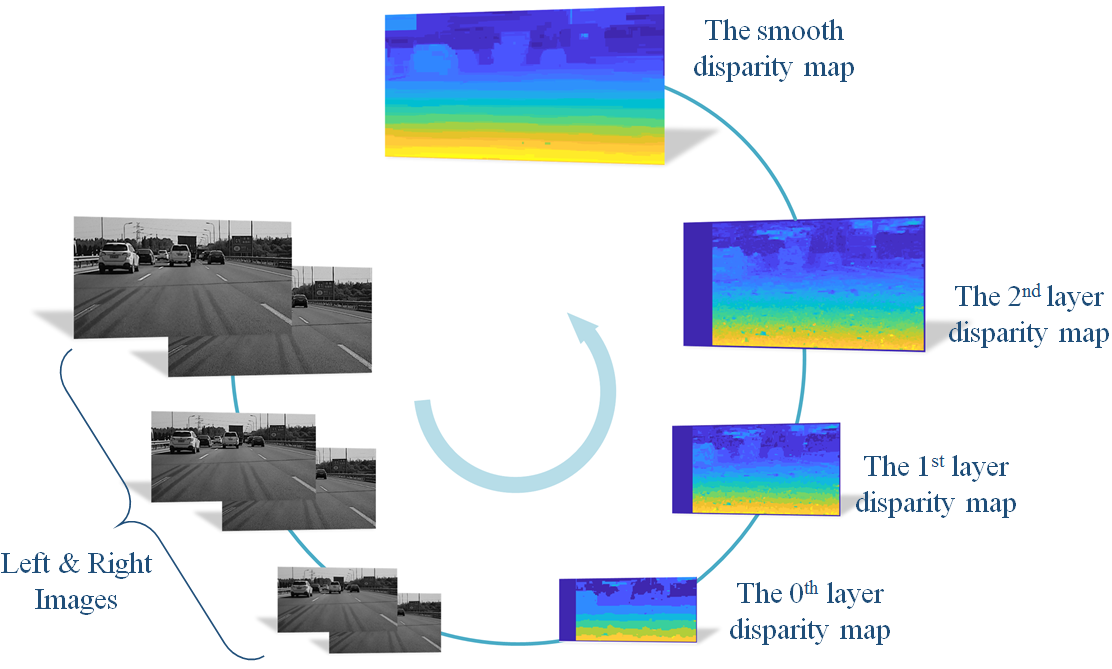}
	\caption{Multi-scale disparity map.}
	\label{mutil_scale_disp_map}
\end{figure}

The result shows that our fast binocular stereo matching algorithm is very reliable. Furthermore, we show the disparity maps in each scale on the image pyramid in Figure \ref{mutil_scale_disp_map}. It gives a help to explain how our proposed multi-scale fast MPV algorithm works.

\subsection{Evaluate of System Detection Accuracy and Running Time}
\begin{figure}[!h]
	\centering
	\includegraphics[width=12cm]{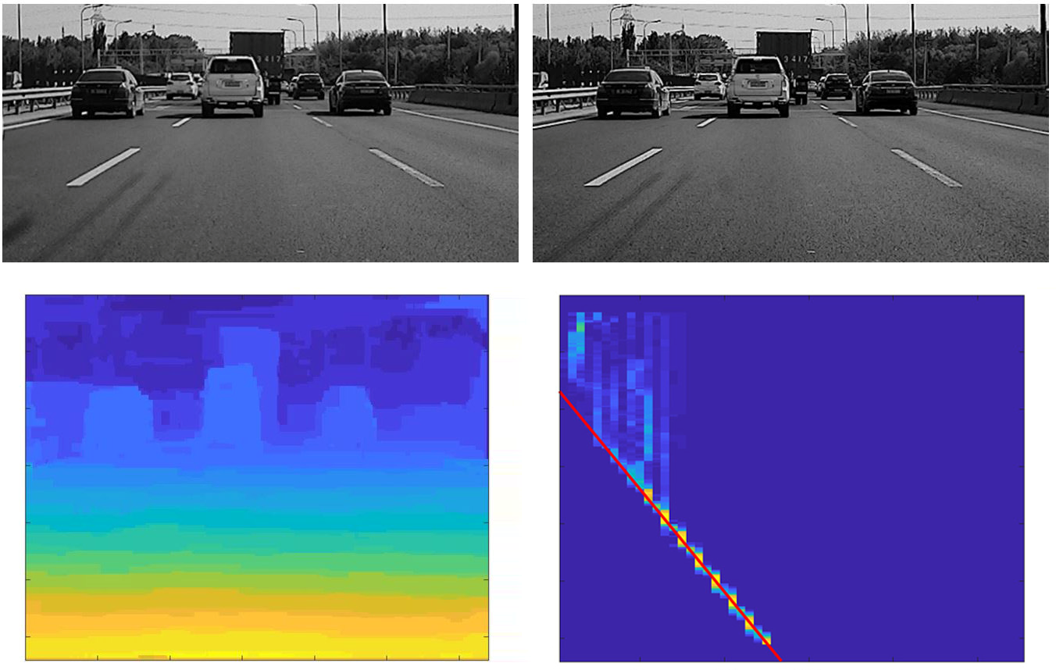}
	\caption{Top: Left image \& Right image; Bottom: Disparity \& Road model.}
	\label{detecte_result}
\end{figure}

We further evaluate our performance of the whole system on our own collected real-time driving datasets. We focus on the evaluation of the detection accuracy and running time. We apply our Fast-MPV  method to the images captured by our vehicle and obtain the disparity map through stereo matching. Then, the road model could be calculated by the disparity map. By comparing the road model and pixels in the disparity map, we can obtain the obstacle regions. In Figure \ref{detecte_result}, the two upper pictures are binocular original images, the lower left picture is the disparity map, and the lower right picture is the calculated road model.

\begin{figure}[!h]
	\centering
	\includegraphics[width=12cm]{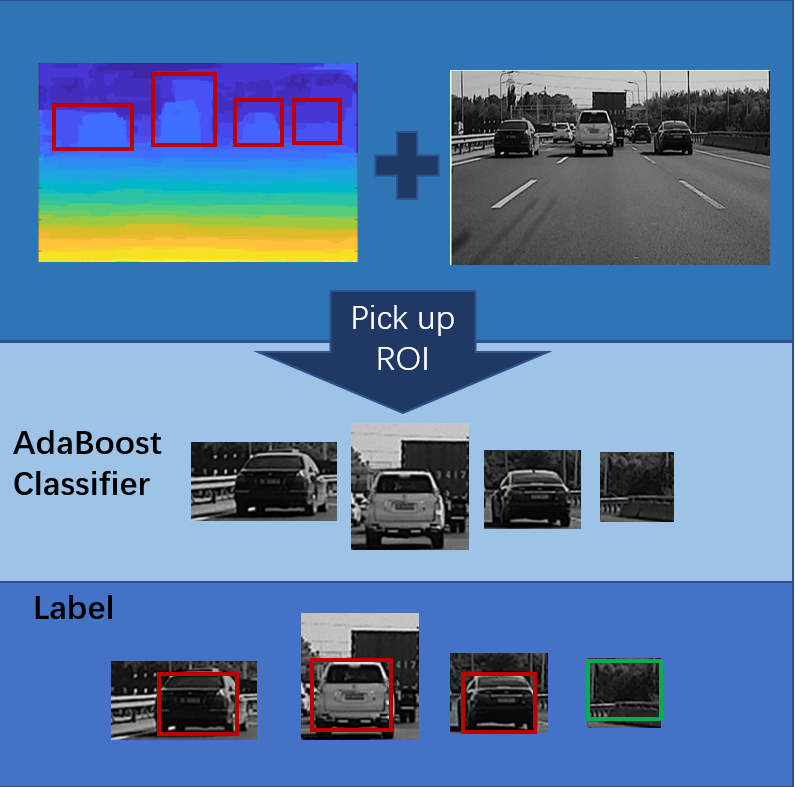}
	\caption{Target recognition flow.}
	\label{target_recognition_flow}
\end{figure}

Combined with the road model and disparity map, we can obtain a binary obstacle ROI image. We use it as a mask to segment the left-eye image. As a result, we can get the fusion map.
As shown in Figure \ref{target_recognition_flow}, every ROI region is treated as a screening window to pick up the ROI gray information as an ROI image.

We collect a large number of samples as ROI images and label them by hand. The cascade AdaBoost classifier is trained by different \emph{Stages} and \emph{maxDepth}, in which \emph{Stages} indicate the series of classifiers and \emph{maxDepth} is the maximum depth of a weak classifier tree. We use the Gentle AdaBoost. The maximum error detection rate for each level of the classifier is $0.5$. Therefore, we can obtain a trained AdaBoost classifier to detect the ROI image. This classifier can recognize whether there is an obstacle in the ROI image. During the experiments, we have the following findings.

\begin{figure}[!h]
	\centering
	\includegraphics[width=16cm]{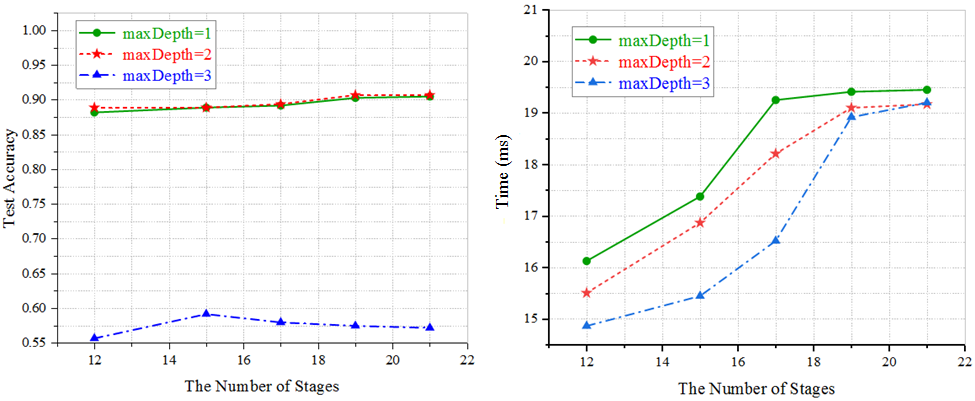}
	\caption{AdaBoost results of maxDepth \& stages. Left: Stages-Accuracy; Right: Stages-Time.}
	\label{Fig27ab}
\end{figure}

\begin{figure}[!h]
	\centering
	\includegraphics[width=8cm]{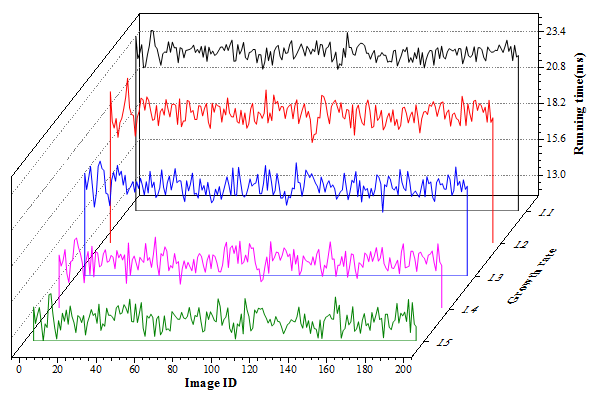}
    \includegraphics[width=8cm]{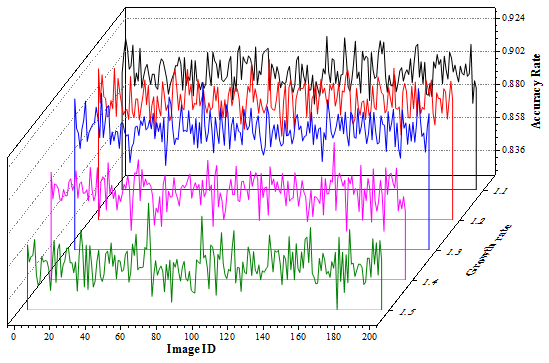}
	\caption{AdaBoost results of growth rate \& window size. Left: Growth rate-Time; Right: Growth rate-Accuracy.}
	\label{Fig28ab}
\end{figure}

\begin{itemize}
  \item As the stages increase, the detection accuracy is not significantly improved, but the detection time is expanded. It means that the number of \emph{stages} is not the key fact affecting the detection accuracy of our classifier with the LBP characteristics. Besides, excessive \emph{stages} will extend the detection time.
  \item As the \emph{maxDepth} increases, the accuracy does not increase. In Figure \ref {Fig27ab}, the classifier is overfitted at \emph{maxDepth} being $3$. At the same time, there is no significant difference in the accuracy of the test between \emph{maxDepth} $1$ and \emph{maxDepth} $2$. However, for the same \emph{stage}, the test time is extended with the increase of \emph{maxDepth}.
  \item Both the size and the growth rate of the detection window affect the detection time. Because we aim at obstacles $10$ meters away, those obstacles generally range in size from $30 \times 30$ to $90 \times 90$ in the image. In addition, we set the growth rate being $1.3$, which means that there are $4$ cycles of detection from the minimum window to the maximum window. As shown in Figure \ref{Fig28ab}, this parameter setting is designed to ensure real-time requirements and greater accuracy of detection.
  \item In order to ensure the detection effect and real-time requirements, we decide to set the stages as $17$ and \emph{maxDepth} as $2$. The accuracy is about $90\%$ and the process time is about $18.2$ms.
\end{itemize}

\begin{figure}[!h]
	\centering
	\includegraphics[width=10cm]{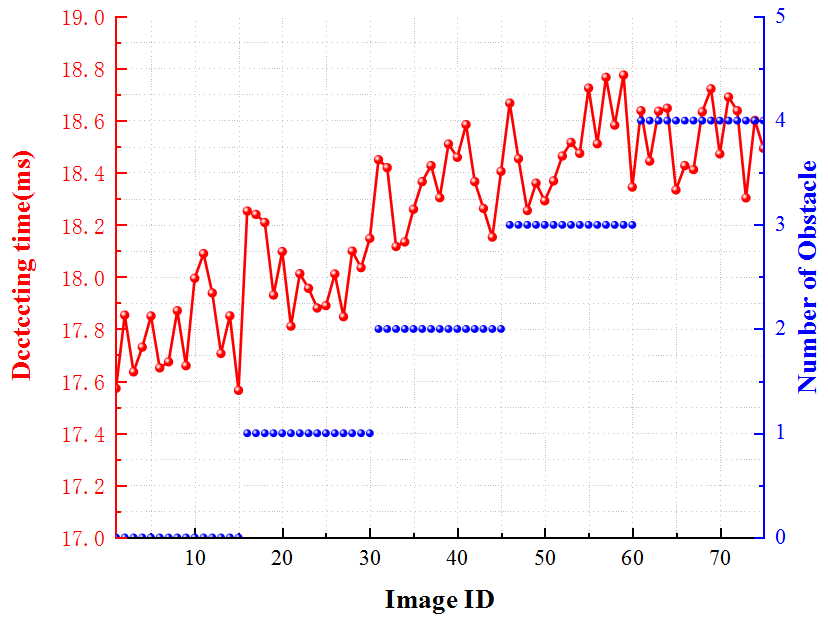}
	\caption{Detection time (There are five comparison groups that have 0, 1, 2, 3 and 4 obstacles, respectively. Each group is contained by 15 images.).}
	\label{detection_time}
\end{figure}

There is only a small difference in time when detecting different numbers of obstacles in our algorithm, as shown in Figure \ref{detection_time}. Since our method is used for autonomous vehicles which takes notice of pedestrians, vehicles and so on, our system focuses on object detection of close obstacles. In addition, we limit the size of the obstacle pixels. Therefore, we can detect $4$ obstacles almost at the same time. Our approach is more robust in terms of the time it takes to detect multiple obstacles.

\begin{table}
	\centering	
	\caption{Results of The Running Time.}
	\begin{tabular}[!h]{c|c}		
		\toprule
		Algorithm & Running time(ms) \\
		\midrule
		UV-disparity \cite{gao2011obstacle} & $47$ \\
        Symmetry \cite{li2002study} & $42$ \\
        AdaBoost & $35$ \\
        Our method (Haar) & $\textbf{18.3}$ \\
        Our method (LBP) & $\textbf{18.2}$\\
		\bottomrule	
	\end{tabular}
    \label{running_time}
\end{table}

\begin{figure}[!h]
	\centering
	\includegraphics[width=14cm]{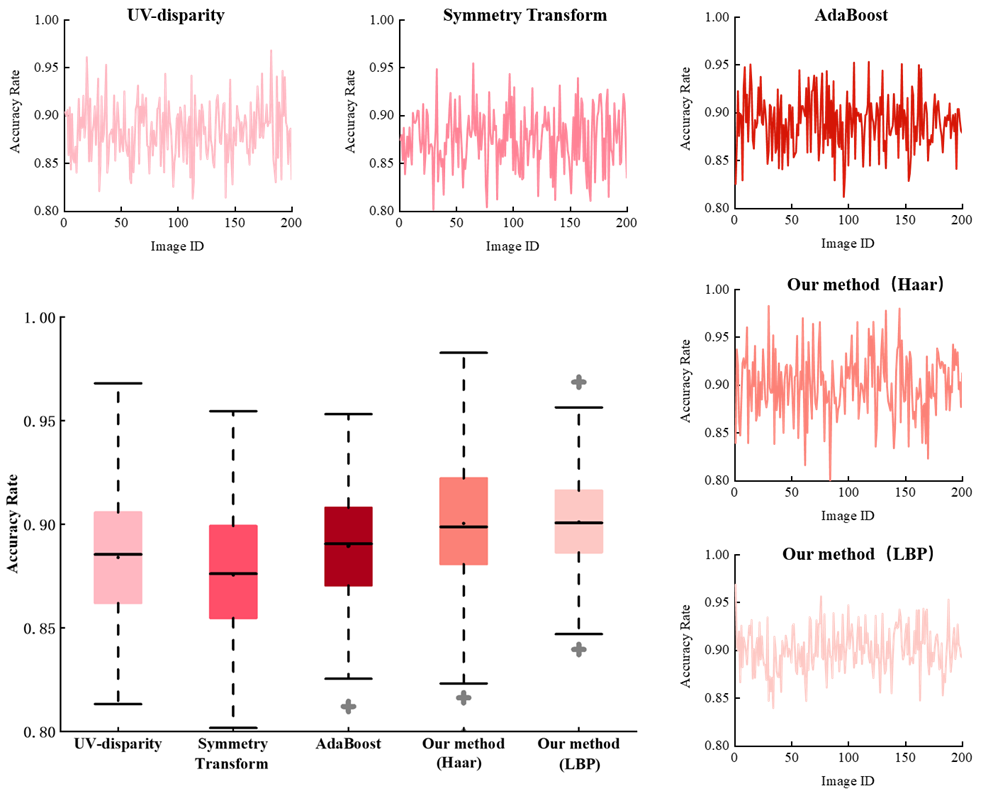}
	\caption{Different algorithm results.}
	\label{diff_alg_result}
\end{figure}

Table \ref{running_time} illustrates the comparison results of the running times among our methods and some other algorithms. The running time in our experiment is calculated by detecting the average of $200$ images. The selected comparison algorithms are typical obstacle detection algorithms for intelligent vehicles. We can see that though the detection accuracies are close, as shown in Figure \ref{diff_alg_result}, our approach has obvious advantage in terms of running time.

The running times of our system with Haar-like features or LBP features are almost the same. However, we found that using the LBP feature detection is more accurate and stable than using the Haar-like feature. Therefore, the LBP feature is more powerful than the Haar feature.

\subsection{Evaluate Detection accuracy under different weather and lighting condition}
\begin{figure}[!h]
	\centering
	\includegraphics[width=10cm]{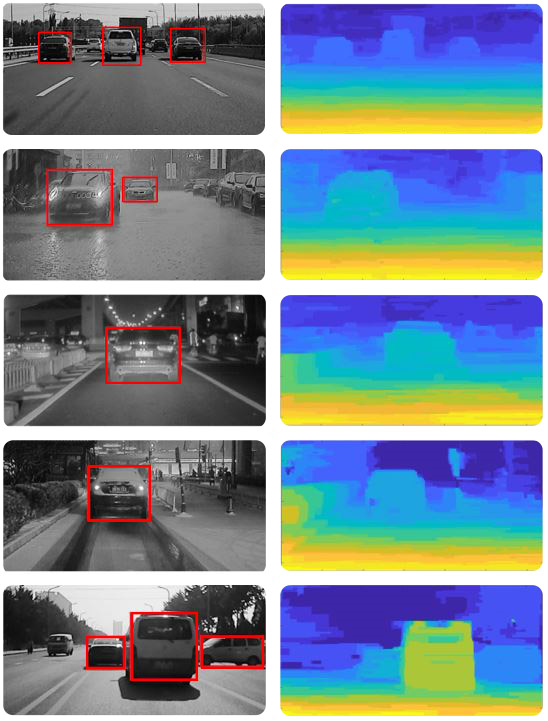}
	\caption{The results under different weather and lighting conditions.}
	\label{weather}
\end{figure}

Different weather and light conditions have different effects on the detection accuracy of the system. The reason is that they have heavy effects on the disparity maps. Figure \ref{weather} illustrates the detection results (left images) and their corresponding disparity maps (right images). From top to bottom, we show the detection results under different weather conditions, such as sunny, rainy, night, snowy and backlight. The experiment results demonstrate that our system is robust in various scenarios.

\begin{figure}[!h]
	\centering
	\includegraphics[width=14cm]{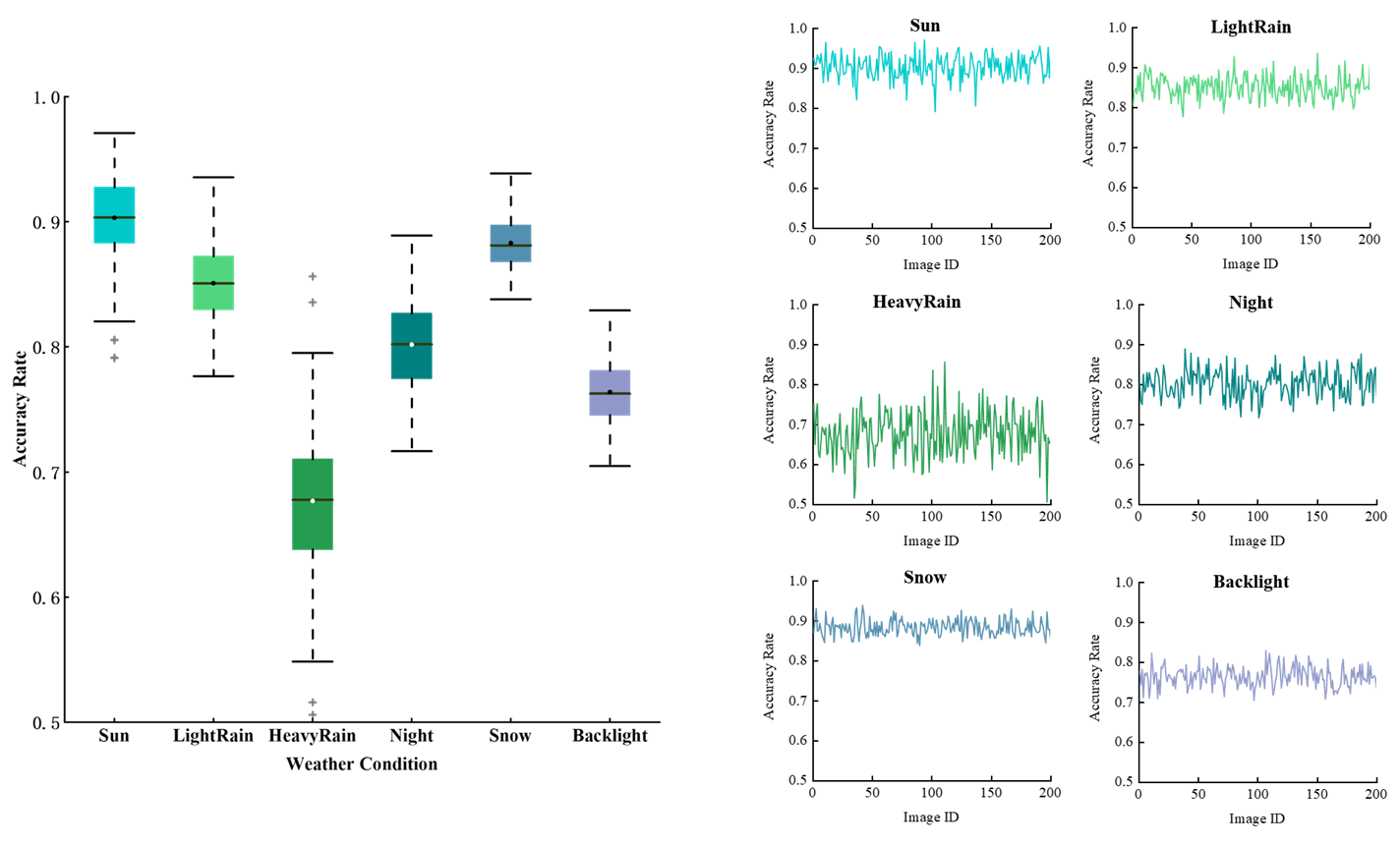}
	\caption{The results of the accuracy rate under different weather and lighting conditions.}
	\label{weather_lighting}
\end{figure}

Rainy days and nights are two main scenarios that reduce detection accuracy. Due to the poor lighting conditions at night, image quality is expected to decrease, however, the experiment results show that our stereo matching algorithm is still robust enough to detect the objects. Low brightness will definitely have a slight effect on the edge of the object recognition, which may cause the classifier to reduce the positional accuracy. However, our system also shows good performance with fill light. On rainy days, the image is blurred due to the occlusion of the rain. In the heavy rain, the accuracy of our system can be greatly reduced. Snowy test data shows that the test results are second excellent. This is mainly because the LBP features have the characteristics of gray invariance. The result of the backlight test data is due to the unclear edge of the object; however, our stereo matching algorithm provides stable positioning results. The corresponding experiment results are shown in Figure \ref{weather_lighting} and Table \ref{accuracy_rate}, respectively, in which the statistical results of $200$ images in each group are demonstrated.

\begin{table}
	\centering	
	\caption{Result of Accuracy Rate.}
	\begin{tabular}[!h]{c|c}		
		\toprule
		Weather Condition & Accuracy Rate \\
		\midrule
		Sunny & $0.905$ \\
        Light Rain & $0.851$ \\
        Heavry Rain & $0.687$ \\
        Night & $0.806$ \\
        Snowy & $0.882$ \\
        Backlight & $0.763$ \\
		\bottomrule	
	\end{tabular}
    \label{accuracy_rate}
\end{table}

\begin{figure}
\centering
{\includegraphics[width=16cm]{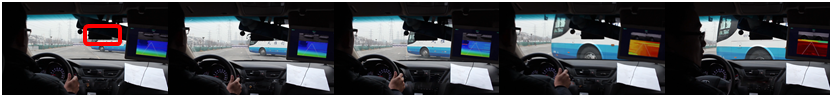}} \newline
{(a) \textbf{Bus body} test from $30$m to $3$m. Object is detected in $1^{st}$ image and alarm is active is $2^{nd}$ image}

{\includegraphics[width=16cm]{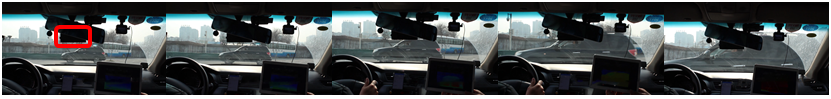}} \newline
{(b) \textbf{Car body} test from $30$m to $3$m. Object is detected in $1^{st}$ image and alarm is active is $2^{nd}$ image}

{\includegraphics[width=16cm]{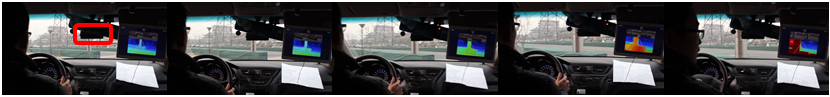}} \newline
{(c) \textbf{Green belt} test from $20$m to $3$m. Object is detected in $1^{st}$ image and alarm is active is $2^{nd}$ image}

{\includegraphics[width=16cm]{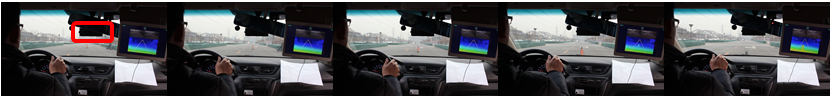}} \newline
{(d) \textbf{Traffic cone} test from $25$m to $3$m. Object is detected in $2^{nd}$ image and alarm is active is $3^{rd}$ image}

{\includegraphics[width=16cm]{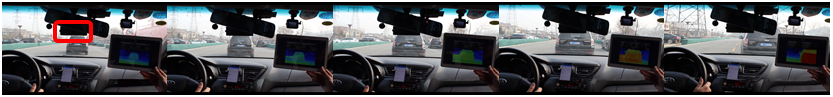}} \newline
{(e) \textbf{Car rear} test from $20$m to $3$m. Object is detected in $1^{st}$ image and alarm is active is $2^{nd}$ image}

\caption{System test. From top to bottom: bus body, car body, green belt, traffic cone, car rear.}
\label{system_test}
\end{figure}

As a system built on a mobile platform, it can be easily carried in a variety of driving environments to achieve ADAS functions. We show the actual effect of the system when it is installed on a small passenger car. As mentioned above, we install the equipment in the middle of the windshield of this passenger car, where our system is in the red box in Figure \ref{system_test}.

In system test, we choose different scenarios to test the obstacle detection function of the system, including the bus body, car body, green belt, traffic cone and car rear. The screen on the right is an external display device, not part of the system itself, just to show our stereo matching effect in the experiment. In the experiment, the driver travels from a distance to the obstacle (from left to right in Figure \ref{system_test}, the test system stably detects the obstacles, and sends the alarms at the preset alarm distance.

\begin{figure}[!h]
	\centering
	\includegraphics[width=12cm]{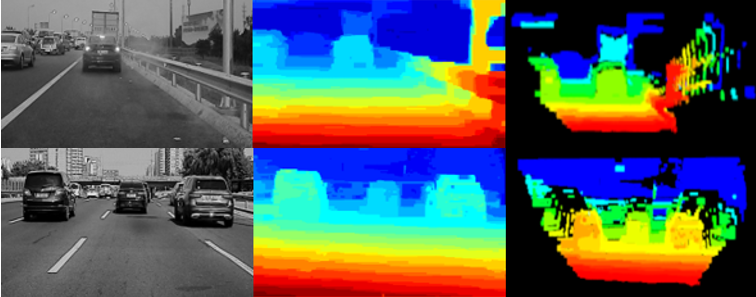}
	\caption{Pseudo-LiDAR signal vs visual disparity map. Top: the evening scene in the rain. Bottom: Sunny afternoon scenes. Left: grey image on binocular left view. Middle: disparity map. Right: pseudo-LiDAR points.}
	\label{cloud_points}
\end{figure}

The test results show that the system is applicable not only for detecting standard obstacles, such as the rear of the car, car body, etc., but also for detecting non-standard obstacles, such as green belt, traffic cones, etc. The test results of different weather conditions and test scenarios show that our system is fully competent for the functional requirements of automatic driving ADAS in the complex environment. Besides, based on our dense disparity map, we can also calculate dense point cloud maps, as shown in Figure \ref{cloud_points}. That can be considered as pseudo-LiDAR representations \cite{wang2019pseudo}.

\subsection{Evaluation of System Module Performance}
At last, the system hardware performance is tested. The influence of the frame rate of image acquisition on the system is shown in Table \ref{fps}. With the increase of image acquisition frame rate, the CPU occupancy rate and image processing frame rate increase significantly, but the growth rate of processing frame rate is lower than the acquisition frame rate, and the memory growth is not obvious. According to this experiment, we set the acquisition frame rate to $12.5$fps, while the processing frame rate is about equal to the same frame rate, and the data flow is most efficient.

\begin{table}
	\centering	
	\caption{The Influence of Frame Rate.}
	\begin{tabular}[!h]{c|ccc}		
		\toprule
		Frame Rate(fps) & $10$ & $15$ & $20$ \\
		\midrule
		CPU Occupancy Rate & $31.25\%$ & $42.75\%$ & $48\%$ \\
        Memory Usage & $54$M & $54$M & $55$M \\
        Processing Frame Rate & $9.7$ & $12.9$ & $13.1$ \\
		\bottomrule	
	\end{tabular}
    \label{fps}
\end{table}

The effect of ambient temperature is shown in Table \ref{temperature}. Our system can operate at ambient temperature of $20$ to $65^\circ$C and the power of full load is not more than $10$W. In order to protect the chip, while the chip temperature exceeds $95^\circ$C, the system will automatically power off.

\begin{table}
	\centering	
	\caption{The Effect of ambient temperature.}
	\begin{tabular}[!h]{c|cccc}		
		\toprule
		Ambient Temperature ($^\circ$C) & $20$ & $55$ & $60$ & $65$ \\
		\midrule
		Chip Temperature ($^\circ$C) & $45 \sim 50$ & $70 \sim 80$ & $80 \sim 85$ & $85 \sim 90$ \\
        Full Load Power (W) & $10$ & $10$ & $10$ & $10$ \\
        Standby Power (W) & $3$ & $3$ & $3$ & $3$ \\
		\bottomrule	
	\end{tabular}
    \label{temperature}
\end{table}

The running time of main task modules under different GPU frequencies is shown in Table \ref{gpu_frequencies}. Running time shows that the operation efficiency of different task modules under different GPU frequencies. Since the calibration and obstacle extraction task do not run at GPU, their running time do not change much. With the increase of GPU frequency, the running time of stereo matching task module decreases and processing frame rate increases. At the same GPU frequency, the stereo matching task's efficiency is also related to image acquisition frame rate. In general, when the frame rate is consistent, the higher the frequency, the shorter the processing time; when the frequency is consistent, the higher the frame rate, the shorter the processing time.

\begin{table}
	\centering	
	\caption{System Performance at Different GPU Frequencies.}
	\begin{tabular}[!h]{c|ccccc}
    \toprule
    Task Module & \multicolumn{5}{c}{Running Time(ms)} \\
    \midrule
    GPU Frequencies & \tabincell{c}{324MHz\\(24fps)} & \tabincell{c}{648MHz\\(24fps)} & \tabincell{c}{852MHz\\(24fps)} & \tabincell{c}{324MHz\\(10fps)} & \tabincell{c}{Auto\\(20fps)} \\
    \midrule
    \multirow{3}{*}{\tabincell{c}{Calibration \\Task (CPU)}} & $31$ & $27$ & $27$ & $38$ & $32$ \\
    & $31$ & $28$ & $26$ & $38$ & $32$ \\
    & $31$ & $29$ & $28$ & $39$ & $34$ \\
    \midrule
    \multirow{3}{*}{\tabincell{c}{Stereo Matching \\ Task (GPU)}} & $96$ & $58$ & $55$ & $102$ & $77$ \\
    & $93$ & $57$ & $50$ & $100$ & $73$ \\
    & $96$ & $59$ & $52$ & $103$ & $81$ \\
    \midrule
    \multirow{3}{*}{\tabincell{c}{Obstacle Extraction \\ Task (CPU)}} & $56$ & $50$ & $50$ & $67$ & $57$ \\
    & $55$ & $47$ & $46$ & $63$ & $56$ \\
    & $53$ & $48$ & $43$ & $61$ & $52$ \\
    \bottomrule	
	\end{tabular}
    \label{gpu_frequencies}
\end{table}

We also tested the performance of obstacle extraction under different frame rates by setting the ARM being $4$ CPUs and frequency being $1.2$G. It is shown in Table \ref{oem}. When the processing frame rate is $10$fps, the mean CPU occupancy rate of the obstacle extraction module is $17.55\%$ and memory usage is $0.1\%$; while the processing frame rate is $15$fps, the mean CPU occupancy rate and memory usage are $16.86\%$ and $0.14$, respectively.

\begin{table}
	\centering	
	\caption{Obstacle Extraction Module.}
	\begin{tabular}[!h]{c|c|c|c|c}
    \toprule
    \multirow{2}{*}{} & \multicolumn{2}{c|}{10fps} & \multicolumn{2}{c}{15fps} \\
    \cline{2-5} & \tabincell{c}{CPU occupancy \\ rate (\%)} & \tabincell{c}{Memory usage\\(\%)} & \tabincell{c}{CPU occupancy \\ rate (\%)} & \tabincell{c}{Memory usage \\ (\%)} \\
    \midrule
    \multirow{5}{*}{\tabincell{c}{Obstacle \\ Extraction \\Module}} & 17.73 & 0.1 & 13.30 & 0.1 \\
    & 18.25 & 0.10 & 17.13 & 0.20 \\
    & 16.78 & 0.10 & 18.05 & 0.30 \\
    & 17.65 & 0.30 & 19.55 & 0.10 \\
    & 17.33 & 0.01 & 16.28 & 0.01 \\
    \midrule
    Mean Value & 17.85 & 0.12 & 16.86 & 0.14 \\
    \bottomrule	
	\end{tabular}
    \label{oem}
\end{table}

\section{Conclusion}
A new robust real-time advanced driver assistance system based on mobile platform is proposed in this paper, which can be applied directly to intelligent driving. There are four major innovations in the system. First, a stereo calibration system is built, which can automatically implement fast calibration for    binocular camera. Secondly, a multi-scale fast MPV algorithm is proposed. It can provide the dense disparity information in real-time for intelligent vehicles. Thirdly, a superior performance cascade AdaBoost classifier is trained, which can provide target detection and recognition in real-time. Fourthly, the distributed computing method and efficient data management approach is advanced, which further improves the performance of the system. The extensive experiment results show that our system not only improves the recognition rate on benchmark database, but also has the applicability in the field of commercial real-time intelligent driving. Our future work will focus on the accuracy improvement under extreme weather conditions.

\section{Acknowledgment}
This work was supported by the National Science Foundation of China under Grant 61673381, Multi-year research grant of University of Macau MYRG2017-00218-FST and MYRG2018-00111-FST.

\bibliographystyle{unsrt}
%\bibliography{myrefs}

\begin{thebibliography}{10}

\bibitem{zhu2017overview}
Hao Zhu, Ka-Veng Yuen, Lyudmila Mihaylova, and Henry Leung.
\newblock Overview of environment perception for intelligent vehicles.
\newblock {\em IEEE Transactions on Intelligent Transportation Systems},
  18(10):2584--2601, 2017.

\bibitem{long2014real}
Qian Long, Qiwei Xie, Seiichi Mita, Kazuhisa Ishimaru, and Noriaki Shirai.
\newblock A real-time dense stereo matching method for critical environment
  sensing in autonomous driving.
\newblock In {\em 17th International IEEE Conference on Intelligent
  Transportation Systems (ITSC)}, pages 853--860. IEEE, 2014.

\bibitem{hata2015feature}
Alberto~Y Hata and Denis~F Wolf.
\newblock Feature detection for vehicle localization in urban environments
  using a multilayer lidar.
\newblock {\em IEEE Transactions on Intelligent Transportation Systems},
  17(2):420--429, 2015.

\bibitem{xie2018pixels}
Shichao Xie, Diange Yang, Kun Jiang, and Yuanxin Zhong.
\newblock Pixels and 3-d points alignment method for the fusion of camera and
  lidar data.
\newblock {\em IEEE Transactions on Instrumentation and Measurement}, 2018.

\bibitem{gao2018object}
Hongbo Gao, Bo~Cheng, Jianqiang Wang, Keqiang Li, Jianhui Zhao, and Deyi Li.
\newblock Object classification using cnn-based fusion of vision and lidar in
  autonomous vehicle environment.
\newblock {\em IEEE Transactions on Industrial Informatics}, 14(9):4224--4231,
  2018.

\bibitem{zhao2018key}
Jianfeng Zhao, Bodong Liang, and Qiuxia Chen.
\newblock The key technology toward the self-driving car.
\newblock {\em International Journal of Intelligent Unmanned Systems},
  6(1):2--20, 2018.

\bibitem{van2018autonomous}
Jessica Van~Brummelen, Marie O¨Brien, Dominique Gruyer, and Homayoun
  Najjaran.
\newblock Autonomous vehicle perception: The technology of today and tomorrow.
\newblock {\em Transportation research part C: emerging technologies},
  89:384--406, 2018.

\bibitem{zhang2016study}
Xinyu Zhang, Hongbo Gao, Mu~Guo, Guopeng Li, Yuchao Liu, and Deyi Li.
\newblock A study on key technologies of unmanned driving.
\newblock {\em CAAI Transactions on Intelligence Technology}, 1(1):4--13, 2016.

\bibitem{montemerlo2008junior}
Michael Montemerlo, Jan Becker, Suhrid Bhat, Hendrik Dahlkamp, Dmitri Dolgov,
  Scott Ettinger, Dirk Haehnel, Tim Hilden, Gabe Hoffmann, Burkhard Huhnke,
  et~al.
\newblock Junior: The stanford entry in the urban challenge.
\newblock {\em Journal of field Robotics}, 25(9):569--597, 2008.

\bibitem{kammel2008team}
S{\"o}ren Kammel, Julius Ziegler, Benjamin Pitzer, Moritz Werling, Tobias
  Gindele, Daniel Jagzent, Joachim Schr{\"o}der, Michael Thuy, Matthias Goebl,
  Felix~von Hundelshausen, et~al.
\newblock Team annieway's autonomous system for the 2007 darpa urban challenge.
\newblock {\em Journal of Field Robotics}, 25(9):615--639, 2008.

\bibitem{weng1992camera}
Juyang Weng, Paul Cohen, and Marc Herniou.
\newblock Camera calibration with distortion models and accuracy evaluation.
\newblock {\em IEEE Transactions on Pattern Analysis \& Machine Intelligence},
  (10):965--980, 1992.

\bibitem{scharstein2002taxonomy}
Daniel Scharstein and Richard Szeliski.
\newblock A taxonomy and evaluation of dense two-frame stereo correspondence
  algorithms.
\newblock {\em International journal of computer vision}, 47(1-3):7--42, 2002.

\bibitem{zhang2000flexible}
Zhengyou Zhang.
\newblock A flexible new technique for camera calibration.
\newblock {\em IEEE Transactions on pattern analysis and machine intelligence},
  22, 2000.

\bibitem{smith2005automatic}
Lyndon~N Smith and Melvyn~L Smith.
\newblock Automatic machine vision calibration using statistical and neural
  network methods.
\newblock {\em Image and Vision Computing}, 23(10):887--899, 2005.

\bibitem{zhao2011camera}
Ping Zhao, Yong-kui Li, Li-jun Chen, and Xue-wei Bai.
\newblock Camera calibration technology based on circular points for binocular
  stereovision system.
\newblock In {\em International Workshop on Computer Science for Environmental
  Engineering and EcoInformatics}, pages 356--363. Springer, 2011.

\bibitem{dey2006richardson}
Nicolas Dey, Laure Blanc-Feraud, Christophe Zimmer, Pascal Roux, Zvi Kam,
  Jean-Christophe Olivo-Marin, and Josiane Zerubia.
\newblock Richardson--lucy algorithm with total variation regularization for 3d
  confocal microscope deconvolution.
\newblock {\em Microscopy research and technique}, 69(4):260--266, 2006.

\bibitem{kong2010general}
Hui Kong, Jean-Yves Audibert, and Jean Ponce.
\newblock General road detection from a single image.
\newblock {\em IEEE Transactions on Image Processing}, 19(8):2211--2220, 2010.

\bibitem{guo2012robust}
Chunzhao Guo, Seiichi Mita, and David McAllester.
\newblock Robust road detection and tracking in challenging scenarios based on
  markov random fields with unsupervised learning.
\newblock {\em IEEE Transactions on intelligent transportation systems},
  13(3):1338--1354, 2012.

\bibitem{hu2005uv}
Zhencheng Hu and Keiichi Uchimura.
\newblock Uv-disparity: an efficient algorithm for stereovision based scene
  analysis.
\newblock In {\em IEEE Proceedings. Intelligent Vehicles Symposium, 2005.},
  pages 48--54. IEEE, 2005.

\bibitem{nedevschi2004high}
Sergiu Nedevschi, Radu Danescu, Dan Frentiu, Tiberiu Marita, Florin Oniga,
  Ciprian Pocol, Thorsten Graf, and Rolf Schmidt.
\newblock High accuracy stereovision approach for obstacle detection on
  non-planar roads.
\newblock {\em Proc. IEEE INES}, pages 211--216, 2004.

\bibitem{sappa2007road}
Angel~D Sappa, Rosa Herrero, Fadi Dornaika, David Ger{\'o}nimo, and Antonio
  L{\'o}pez.
\newblock Road approximation in euclidean and v-disparity space: a comparative
  study.
\newblock In {\em International Conference on Computer Aided Systems Theory},
  pages 1105--1112. Springer, 2007.

\bibitem{forney1973viterbi}
G~David Forney.
\newblock The viterbi algorithm.
\newblock {\em Proceedings of the IEEE}, 61(3):268--278, 1973.

\bibitem{xie2017integration}
Qiwei Xie, Qian Long, and Seiichi Mita.
\newblock Integration of optical flow and multi-path-viterbi algorithm for
  stereo vision.
\newblock {\em International Journal of Wavelets, Multiresolution and
  Information Processing}, 15(03):1750022, 2017.

\bibitem{power2013heterogeneous}
Jason Power, Arkaprava Basu, Junli Gu, Sooraj Puthoor, Bradford~M Beckmann,
  Mark~D Hill, Steven~K Reinhardt, and David~A Wood.
\newblock Heterogeneous system coherence for integrated cpu-gpu systems.
\newblock In {\em Proceedings of the 46th Annual IEEE/ACM International
  Symposium on Microarchitecture}, pages 457--467. ACM, 2013.

\bibitem{fleuret2004fast}
Fran{\c{c}}ois Fleuret.
\newblock Fast binary feature selection with conditional mutual information.
\newblock {\em Journal of Machine learning research}, 5(Nov):1531--1555, 2004.

\bibitem{vineet2008cuda}
Vibhav Vineet and PJ~Narayanan.
\newblock Cuda cuts: Fast graph cuts on the gpu.
\newblock In {\em 2008 IEEE Computer Society Conference on Computer Vision and
  Pattern Recognition Workshops}, pages 1--8. IEEE, 2008.

\bibitem{kirk2007nvidia}
David Kirk et~al.
\newblock Nvidia cuda software and gpu parallel computing architecture.
\newblock In {\em ISMM}, volume~7, pages 103--104, 2007.

\bibitem{bahl1974optimal}
Lalit Bahl, John Cocke, Frederick Jelinek, and Josef Raviv.
\newblock Optimal decoding of linear codes for minimizing symbol error rate
  (corresp.).
\newblock {\em IEEE Transactions on information theory}, 20(2):284--287, 1974.

\bibitem{chambolle2004algorithm}
Antonin Chambolle.
\newblock An algorithm for total variation minimization and applications.
\newblock {\em Journal of Mathematical imaging and vision}, 20(1-2):89--97,
  2004.

\bibitem{ranftl2013minimizing}
Rene Ranftl, Thomas Pock, and Horst Bischof.
\newblock Minimizing tgv-based variational models with non-convex data terms.
\newblock In {\em International Conference on Scale Space and Variational
  Methods in Computer Vision}, pages 282--293. Springer, 2013.

\bibitem{birchfield1999multiway}
Stan Birchfield and Carlo Tomasi.
\newblock Multiway cut for stereo and motion with slanted surfaces.
\newblock In {\em Proceedings of the seventh IEEE international conference on
  computer vision}, volume~1, pages 489--495. IEEE, 1999.

\bibitem{liu2011objective}
Zheng Liu, Erik Blasch, Zhiyun Xue, Jiying Zhao, Robert Laganiere, and Wei Wu.
\newblock Objective assessment of multiresolution image fusion algorithms for
  context enhancement in night vision: a comparative study.
\newblock {\em IEEE transactions on pattern analysis and machine intelligence},
  34(1):94--109, 2011.

\bibitem{garcia2017sensor}
Fernando Garcia, David Martin, Arturo De~La~Escalera, and Jose~Maria Armingol.
\newblock Sensor fusion methodology for vehicle detection.
\newblock {\em IEEE Intelligent Transportation Systems Magazine},
  9(1):123--133, 2017.

\bibitem{spacey2012robust}
Simon~A Spacey, Wolfram Wiesemann, Daniel Kuhn, and Wayne Luk.
\newblock Robust software partitioning with multiple instantiation.
\newblock {\em INFORMS Journal on Computing}, 24(3):500--515, 2012.

\bibitem{boschetti2016using}
Marco~A Boschetti, Vittorio Maniezzo, and Francesco Strappaveccia.
\newblock Using gpu computing for solving the two-dimensional guillotine
  cutting problem.
\newblock {\em INFORMS Journal on Computing}, 28(3):540--552, 2016.

\bibitem{date2019level}
Ketan Date and Rakesh Nagi.
\newblock Level 2 reformulation linearization technique--based parallel
  algorithms for solving large quadratic assignment problems on graphics
  processing unit clusters.
\newblock {\em INFORMS Journal on Computing}, 31(4):771--789, 2019.

\bibitem{konolige1998small}
Kurt Konolige.
\newblock Small vision systems: Hardware and implementation.
\newblock In {\em Robotics research}, pages 203--212. Springer, 1998.

\bibitem{li2018binocular}
Tongtong Li, Changying Liu, Yang Liu, Tianhao Wang, and Dapeng Yang.
\newblock Binocular stereo vision calibration based on alternate adjustment
  algorithm.
\newblock {\em Optik}, 173:13--20, 2018.

\bibitem{wang2004image}
Zhou Wang, Alan~C Bovik, Hamid~R Sheikh, Eero~P Simoncelli, et~al.
\newblock Image quality assessment: from error visibility to structural
  similarity.
\newblock {\em IEEE transactions on image processing}, 13(4):600--612, 2004.

\bibitem{son2006stereo}
Tran~Thai Son and Seiichi Mita.
\newblock Stereo matching algorithm using a simplified trellis diagram
  iteratively and bi-directionally.
\newblock {\em IEICE transactions on information and systems}, 89(1):314--325,
  2006.

\bibitem{lowe2004distinctive}
David~G Lowe.
\newblock Distinctive image features from scale-invariant keypoints.
\newblock {\em International journal of computer vision}, 60(2):91--110, 2004.

\bibitem{choi2012environment}
Jaewoong Choi, Junyoung Lee, Dongwook Kim, Giacomo Soprani, Pietro Cerri,
  Alberto Broggi, and Kyongsu Yi.
\newblock Environment-detection-and-mapping algorithm for autonomous driving in
  rural or off-road environment.
\newblock {\em IEEE Transactions on Intelligent Transportation Systems},
  13(2):974--982, 2012.

\bibitem{caraffi2007off}
Claudio Caraffi, Stefano Cattani, and Paolo Grisleri.
\newblock Off-road path and obstacle detection using decision networks and
  stereo vision.
\newblock {\em IEEE Transactions on Intelligent Transportation Systems},
  8(4):607--618, 2007.

\bibitem{altun2017road}
Melih Altun and Mehmet Celenk.
\newblock Road scene content analysis for driver assistance and autonomous
  driving.
\newblock {\em IEEE transactions on intelligent transportation systems},
  18(12):3398--3407, 2017.

\bibitem{mammeri2016extending}
Abdelhamid Mammeri, Tianyu Zuo, and Azzedine Boukerche.
\newblock Extending the detection range of vision-based vehicular
  instrumentation.
\newblock {\em IEEE Transactions on Instrumentation and Measurement},
  65(4):856--873, 2016.

\bibitem{yuan2015multisensor}
Jing Yuan, Huan Chen, Fengchi Sun, and Yalou Huang.
\newblock Multisensor information fusion for people tracking with a mobile
  robot: A particle filtering approach.
\newblock {\em IEEE transactions on Instrumentation and Measurement},
  64(9):2427--2442, 2015.

\bibitem{lins2015vision}
Romulo~Gon{\c{c}}alves Lins, Sidney~N Givigi, and Paulo Roberto~Gardel Kurka.
\newblock Vision-based measurement for localization of objects in 3-d for
  robotic applications.
\newblock {\em IEEE Transactions on Instrumentation and Measurement},
  64(11):2950--2958, 2015.

\bibitem{schapire2013explaining}
Robert~E Schapire.
\newblock Explaining adaboost.
\newblock In {\em Empirical inference}, pages 37--52. Springer, 2013.

\bibitem{liao2015fast}
Shengcai Liao, Anil~K Jain, and Stan~Z Li.
\newblock A fast and accurate unconstrained face detector.
\newblock {\em IEEE transactions on pattern analysis and machine intelligence},
  38(2):211--223, 2015.

\bibitem{li2008adaboost}
Xuchun Li, Lei Wang, and Eric Sung.
\newblock Adaboost with svm-based component classifiers.
\newblock {\em Engineering Applications of Artificial Intelligence},
  21(5):785--795, 2008.

\bibitem{viola2002fast}
Paul Viola and Michael Jones.
\newblock Fast and robust classification using asymmetric adaboost and a
  detector cascade.
\newblock In {\em Advances in neural information processing systems}, pages
  1311--1318, 2002.

\bibitem{wang2009hog}
Xiaoyu Wang, Tony~X Han, and Shuicheng Yan.
\newblock An hog-lbp human detector with partial occlusion handling.
\newblock In {\em 2009 IEEE 12th international conference on computer vision},
  pages 32--39. IEEE, 2009.

\bibitem{liu2016median}
Li~Liu, Songyang Lao, Paul~W Fieguth, Yulan Guo, Xiaogang Wang, and Matti
  Pietik{\"a}inen.
\newblock Median robust extended local binary pattern for texture
  classification.
\newblock {\em IEEE Transactions on Image Processing}, 25(3):1368--1381, 2016.

\bibitem{aguilar2014robust}
Wilbert~G Aguilar and Cecilio Angulo.
\newblock Robust video stabilization based on motion intention for low-cost
  micro aerial vehicles.
\newblock In {\em 2014 IEEE 11th International Multi-Conference on Systems,
  Signals \& Devices (SSD14)}, pages 1--6. IEEE, 2014.

\bibitem{hu2013online}
Weiming Hu, Jun Gao, Yanguo Wang, Ou~Wu, and Stephen Maybank.
\newblock Online adaboost-based parameterized methods for dynamic distributed
  network intrusion detection.
\newblock {\em IEEE Transactions on Cybernetics}, 44(1):66--82, 2013.

\bibitem{thain2005distributed}
Douglas Thain, Todd Tannenbaum, and Miron Livny.
\newblock Distributed computing in practice: the condor experience.
\newblock {\em Concurrency and computation: practice and experience},
  17(2-4):323--356, 2005.

\bibitem{zhang2012imapreduce}
Yanfeng Zhang, Qixin Gao, Lixin Gao, and Cuirong Wang.
\newblock imapreduce: A distributed computing framework for iterative
  computation.
\newblock {\em Journal of Grid Computing}, 10(1):47--68, 2012.

\bibitem{kshemkalyani2011distributed}
Ajay~D Kshemkalyani and Mukesh Singhal.
\newblock {\em Distributed computing: principles, algorithms, and systems}.
\newblock Cambridge University Press, 2011.

\bibitem{geiger2012we}
Andreas Geiger, Philip Lenz, and Raquel Urtasun.
\newblock Are we ready for autonomous driving? the kitti vision benchmark
  suite.
\newblock In {\em 2012 IEEE Conference on Computer Vision and Pattern
  Recognition}, pages 3354--3361. IEEE, 2012.

\bibitem{hirschmuller2007stereo}
Heiko Hirschmuller.
\newblock Stereo processing by semiglobal matching and mutual information.
\newblock {\em IEEE Transactions on pattern analysis and machine intelligence},
  30(2):328--341, 2007.

\bibitem{geiger2010efficient}
Andreas Geiger, Martin Roser, and Raquel Urtasun.
\newblock Efficient large-scale stereo matching.
\newblock In {\em Asian conference on computer vision}, pages 25--38. Springer,
  2010.

\bibitem{gao2011obstacle}
Yuan Gao, Xiao Ai, John Rarity, and Naim Dahnoun.
\newblock Obstacle detection with 3d camera using uv-disparity.
\newblock In {\em International Workshop on Systems, Signal Processing and
  their Applications, WOSSPA}, pages 239--242. IEEE, 2011.

\bibitem{li2002study}
Bin Li, Rong-ben Wang, and Ke-you Guo.
\newblock Study on machine vision based obstacle detection and recognition
  method for intelligent vehicle.
\newblock {\em Kung lu chiao tung ko chi= journal of highway and transportation
  research and development. Vol. 19, no. 4}, 2002.

\bibitem{wang2019pseudo}
Yan Wang, Wei-Lun Chao, Divyansh Garg, Bharath Hariharan, Mark Campbell, and
  Kilian~Q Weinberger.
\newblock Pseudo-lidar from visual depth estimation: Bridging the gap in 3d
  object detection for autonomous driving.
\newblock In {\em Proceedings of the IEEE Conference on Computer Vision and
  Pattern Recognition}, pages 8445--8453, 2019.

\end{thebibliography}

\end{document}